\documentclass[runningheads]{llncs}
\usepackage[utf8]{inputenc}
\usepackage[noadjust]{cite}
\usepackage{graphicx}

\usepackage{verbatim} 

\usepackage{tikz}
\usetikzlibrary{patterns,shapes.geometric, arrows}
\usepackage{pgfplots}
\DeclareUnicodeCharacter{2212}{−}
\usepgfplotslibrary{groupplots,dateplot}
\pgfplotsset{compat=newest}

\usepackage[title]{appendix}
\usepackage[english]{babel}
\usepackage{todonotes}
\usepackage{hyperref} 
\usepackage{footnote} %
\usepackage{tablefootnote}

\usepackage{cleveref}
\usepackage{xcolor}
\usepackage{chicago}

\usepackage[strings]{underscore}
\DeclareGraphicsExtensions{.jpg,.eps,.pdf,.png} 
\usepackage[font=small,parskip=1pt]{caption} %
\usepackage{subcaption}
    \makeatletter
    \renewcommand*{\@fnsymbol}[1]{\ensuremath{\ifcase#1\or *\or \dagger\or \ddagger\or
        \mathsection\or \mathparagraph\or \|\or **\or \dagger\dagger
        \or \ddagger\ddagger \else\@ctrerr\fi}}
    \makeatother

\begin{document}
\title{An analysis of the transfer learning of convolutional neural networks for artistic images\thanks{Supported by the "IDI 2017" project funded by the IDEX Paris-Saclay, ANR-11-IDEX-0003-02 and T\'el\'ecom Paris.}}
\titlerunning{An analysis of the transfer learning of CNNs for artistic images}
\author{Nicolas Gonthier\inst{1,2}\thanks{Corresponding author} \and
Yann Gousseau\inst{1} \and
Sa\"id Ladjal\inst{1}}
\authorrunning{N. Gonthier et al.}
\institute{LTCI - T\'el\'ecom Paris - Institut Polytechnique de Paris, 91120  Palaiseau, France \and
Universit\'e Paris Saclay, 91190 Saint-Aubin, France \\ 
\email{nicolas.gonthier@telecom-paris.fr}}
\maketitle              %
\begin{abstract}
Transfer learning from huge natural image datasets, fine-tuning of deep neural networks and the use of the corresponding pre-trained networks have become de facto the core of art analysis applications. Nevertheless, the effects of transfer learning are still poorly understood. In this paper, we first use techniques for visualizing the network internal representations in order to provide clues to the understanding of what the network has learned on artistic images. Then, we provide a quantitative analysis of the changes introduced by the learning process thanks to metrics in both the feature and parameter spaces, as well as metrics computed on the set of maximal activation images.
These analyses are performed on several variations of the transfer learning procedure. In particular, we observed that the network could specialize some pre-trained filters to the new image modality and also that higher layers tend to concentrate classes.
Finally, we have shown that a double fine-tuning involving a medium-size artistic dataset can improve the classification on smaller datasets, even when the task changes.
\keywords{Transfer Learning\and Convolutional Neural Network \and Art analysis \and Feature Visualization.}
\end{abstract}

\section{Introduction}
Over the last decade, numerous efforts have been invested in the digitization of fine art, yielding digital collections allowing the preservation and remote access to cultural heritage. %
Such collections, even when available online, can only be fruitfully browsed through the metadata associated with images. In recent years, several research teams have developed search engines dedicated to fine arts for different recognition tasks: Replica \cite{seguin_visual_2016} for visual similarity search or the Oxford Painting Search \cite{crowley_art_2016} for semantics recognition of arbitrary objects. Often, those search engines are based on convolutional neural networks (CNN).
Transfer learning from large-scale natural image datasets such as ImageNet, mostly by fine-tuning large pre-trained networks, has become a de facto standard for art analysis applications. 
Nevertheless, there are large differences in dataset sizes, image style and task specifications between natural images and the target artistic images, and there is little understanding of the effects of transfer learning in this context. In this work, we explore some properties of transfer learning for artistic images, by using both visualization techniques and quantitative studies. 
Visualization techniques permit to understand what the networks have learned on specific artistic datasets, by showing some of their internal representations or giving hints at what aspects of artistic images are important for their understanding.
In particular, we will see that the networks can specify some pre-trained filters in order to adapt them to the new modality of images and also that the network can learn new, highly structured filters specific to artistic images from scratch.
 We also look at the set of the maximal activation images for a given channel to complete our observation.
 
Quantitative results can confirm some intuitive facts about the way networks are modified during fine-tuning.
To quantify the amount of change experienced by networks in different fine-tuning modality, we rely on  feature similarity and the $\ell_2$ distance between models.
We also compute metrics (overlapping ratio and entropy) on the maximal activation images set to this end.
Moreover we experimentally show that fine-tuning first a pretrained ImageNet model on an intermediate artistic dataset may lead to better performance than a direct fine tuning on the target small artistic dataset (for a different tasks). 
Let us emphasize that the goal of this work is not to provide state-of-the-art classification performances, but rather to investigate the way CNNs are modified by classical fine-tuning operations in the specific case of artwork images.

\section{Related Work}
 Our analysis of the adaptation of a deep network to artistic databases uses already well-established tools and methods. In the following we describe these methods and list the relevant related works. 
 
\subsection{Deep Transfer Learning for Art Classification Problems}

  Transfer learning consists in adapting a model trained on a large image database (such as ImageNet \cite{russakovsky_imagenet_2014}) for a new task. This method is the de facto standard when faced with relatively small datasets and has proven its relevance in many works. Two main modalities are possible for transfer learning. The first consists in taking the penultimate output of the pre-trained network to make it the input of a simple classifier \cite{donahue_decaf_2013}. In the following, we refer to this approach as the \emph{off-the-shelf} method. The second option consists in \emph{fine-tuning} (FT) the pre-trained network for the new task \cite{girshick_rich_2014}. One can also argue that the bare architecture of a successful network is in itself a form of transfer learning, as this architecture has proven its relevance to the task of image classification.%

   On bigger datasets, one can fine-tune the weights to adapt the network to a new task. This approach is by far the most used one. For the domain of artistic images, it has been used for style classification~\cite{tan_ceci_2016,lecoutre_recognizing_2017,cetinic_finetuning_2018}, object recognition  in drawings \cite{yin_object_2016} or iconographic characters \cite{madhu_recognizing_2019}, people detection across a variety of artworks \cite{westlake_detecting_2016}, visual link retrieval \cite{seguin_visual_2016}, author classification~\cite{vannoord_learning_2017,sabatelli_deep_2018} or several of those tasks at the same time \cite{bianco_multitask_2019}.
   
  More precisely, \cite{tan_ceci_2016} show that fine-tuning a CNN pretrained on ImageNet outperforms off-the-shelf and training from scratch strategies for style, genre or artist classification.
  In \cite{cetinic_finetuning_2018}, the authors evaluated the impact of domain-specific weight initialization and the kind of fine-tuning used (number of frozen layers for instance) for different art classification tasks. They compared different pre-training with different natural images datasets. They have shown that the bigger (in terms of training images and number of labels), the better will be the results.
  
   A midway strategy between directly fine-tuning a pre-trained network and the mere use of the final network features, when the dataset is small, is to have a two phase fine-tuning, the first one with a relatively large dataset of artworks and  the second on the target dataset. This strategy was shown to be helpful in  \cite{sabatelli_deep_2018}, using the Rijksmuseum dataset for the first fine-tuning. Their findings suggest that the double fine-tuned model focuses more on fine details to perform artist attribution. 
   
    In this work, we will look at the two ways of fine-tuning and the various effects they have on what the network learns to adapt itself to artworks. When using a double fine-tuning, the middle dataset will always be the RASTA dataset (described below). We will also look at the transfer of the bare architecture, which means initializing the weights to random values. Intermediate strategies such as partial freezing of the network will also be studied.

\subsection{Deep Convolutional Neural Network Understanding }
The deep learning community has provided several tools for trying to better understand deep CNNs : feature visualization \cite{erhan_visualizing_2009,olah_feature_2017} and attribution \cite{simonyan_deep_2014}. %
Feature visualization answers questions about what a deep network is responding to in a dataset by generating examples that yield maximum activation. 
Nevertheless, to achieve visually good results and output results that are understandable by humans, it is a necessity to add some constraints \cite{olah_feature_2017} to the optimization, thus avoiding getting adversarial structured noise.
Based on those works, several papers have proposed methodology to determine the way the different features contribute to a classification  \cite{olah_building_2018} by mixing it with attribution methods. Visualization of the optimized images also permits regrouping the filters of the first layers of an InceptionV1 model in some comprehensible groups \cite{olah_overview_2020}.
Such works  \cite{erhan_visualizing_2009,olah_overview_2020,olah_feature_2017} tend to show that a CNN learns meaningful features (for instance eye detector or dog head detector) whereas others show that those networks are  primarily detecting textures \cite{geirhos_imagenettrained_2018}. 
By looking at the channel responses, the authors of \cite{tan_ceci_2016} concluded that lower layers learn simple patterns and higher ones, complex object parts such as portrait shape.
In \cite{strezoski_plugandplay_2017}, the authors look at the feature visualizations and attributions of a small convolutional network  trained on an artistic dataset. Some of the characteristic patterns of the classes (as the circle shape for portrait class) can be found in the visualizations.
In \cite{szabo_visualizing_2020}, the authors visualize the impact of the fine-tuning of a network on fine-grained datasets. They demonstrate various  properties of the transfer learning process such as the speed and characteristics of adaptation, neuron reuse and spatial scale of the represented image features on natural images datasets.
Another way to understand the inner structure of networks is to compute feature similarity between different layers or different models. The recent work \cite{kornblith_similarity_2019} proposes to do this through Centred Kernel Alignement (CKA), a measure that we will use later in this work.

 \subsection{Datasets} 

  Most artistic datasets only contain style or author metadata \cite{lecoutre_recognizing_2017,tan_ceci_2016} instead of depicted objects or iconographic elements. Some datasets are specific to a given class of objects such as person in paintings \cite{westlake_detecting_2016} or to concepts that are specific to art history \cite{cetinic_learning_2019}. %
In~\cite{wilber_bam_2017}, an annotated database of 2.2M contemporary artworks from Behance (website of portfolios from professional artists) is introduced, on which it is shown that fine-tuning improves recognition performances. The OmniArt dataset introduced in \cite{strezoski_omniart_2018} contains 1M historical artworks of 4 different types (from craft to paintings).
Those two large-scale datasets are not openly accessible yet and no models pretrained on them has been shared to the community.
For our experiments we use three datasets which come from different research works. 
The first one contains the largest number of samples and comes from the WikiArt website. It contains 80,000 images tagged with one among 25 artistic styles \cite{lecoutre_recognizing_2017} and is named \emph{RASTA}. Many other works referred to this dataset as the “WikiArt paintings” \cite{tan_ceci_2016} but this variant contains only 25 classes instead of 27.
Due to its size and large diversity, we will mainly use this dataset in the experimental section. 
The second one is the \emph{Paintings} Dataset introduced in \cite{crowley_search_2014}, made of 8629 British painting images with 10 different labels corresponding to common objects.%
 The last dataset is the \emph{IconArt} dataset from \cite{gonthier_weakly_2018} composed of 5955 painting images from Wikicommons with 7 iconographic labels, for instance angel or the crucifixion of Jesus. These two datasets are designed for object classification, similarly to ImageNet.

\section{Analyzing CNNs Trained for Art Classification Tasks} %
In this work, we investigate the effect of fine-tuning in the case of artistic images. In order to do so, we rely both on visualization techniques and quantification of the change the network undergoes. 
Our experimental results are organized in five sections.
 First, we consider an Inception V1 network~\cite{szegedy_going_2015}  pre-trained on ImageNet and fine-tuned on RASTA  for artistic style classification (\cref{subsec:ImageNetToArt}). Then we consider the same architecture with a random initialization (from scratch) trained on RASTA (\cref{subsec:FromScratch}). The \cref{sec:perfo_RASTA,sec:CKA_l2_RASTA} are dedicated to the classification performance and the evaluation of the changes implied by the training on RASTA. 
Finally, we studied the same architecture pre-trained on ImageNet and then fine-tuned first on RASTA and then on a smaller art  dataset for object classification (\cref{subsec:ArtToArt}) to see how using an intermediate art dataset can help.

\paragraph{Feature visualization }
\label{par:featVizu}
The first visualization technique we use consists in generating \emph{optimized images}, as introduced in \cite{olah_feature_2017}. These images are obtained by maximizing the response to a given channel. The entire feature map at a given layer can be decomposed in two spatial dimensions and one dimension depending on the convolutional kernels. A \emph{channel} denotes one element according to this last dimension.
We use the \href{https://github.com/tensorflow/lucid}{Lucid framework} for visualizing convolutional channels via activation maximization. We use Lucid's 2D FFT image representation with decorrelation and 2048 iterations of the gradient ascent.%

\paragraph{Maximal activation images}
We devise another indicator that might be useful for the analysis of the transformation that a network undergoes during its transfer to a different domain. This indicator is the evolution of the \emph{maximal activation images}. For a given channel, we compute the \emph{top 100} images in the target dataset that trigger it the most.
We also compute the information \emph{entropy} over classes for each top 100 images, in order to evaluate the clustering power of the corresponding channel. The entropy is defined as $\frac{1}{maxE} \sum\limits_{classes} -p_c log_2 (p_c)$ with $p_c$ the fraction of images in the top 100 belonging to the class $c$ and $maxE$ the maximal entropy with this number of classes.
Moreover, the top 100 can be computed twice, once at the beginning  and once at the end of the fine-tuning. The percentage of the images that lie in both sets is an indicator of how much the channel has drifted during its adaptation. These percentages are named \emph{overlapping ratio} in the following. They are, in many cases, much higher than what we would expect from a random reshuffling of the dataset. Besides, the combination of this indicator with the visualization technique from \cite{olah_feature_2017} leads to several findings that we will present thereafter. 
\paragraph{Experimental Setup :}
All our visualization experiments use the InceptionV1 \cite{szegedy_going_2015} CNN (also called GoogLeNet). It is a 22 layers with only 7M parameters thanks to the introduction of Inception modules.
The last layer of the network is replaced by a fully connected layer with the number of outputs corresponding to the dataset at hand and where activation function is a softmax for RASTA or a sigmoid for Paintings and IconArt datasets. The loss function is the usual cross-entropy in the first case, and the sum over the classes of binary cross-entropy in the two others. 
The InceptionV1 network is the classical and efficient choice for feature visualization by optimization \cite{olah_feature_2017} although it no longer produces the best classification performances.
We ran experiments with a various number of hyperparameters such as the learning rate for the last layer (classification layer), the learning rate for the transferred layers, the use of a deep supervision, the maximum number of epochs or the possible use of random crops within the input image. The input size of the network is 224 $\times$ 224. For all experiments, we selected the model with the best loss value on the corresponding validation set.

In the following sections, we analyze how the networks have been modified by fine-tuning processes. We present qualitative observations using optimized images and the maximal activation images, as well as quantitative evaluations relying on the $\ell_{2}$ norm of the difference between convolution kernels and the linear CKA measure~\cite{kornblith_similarity_2019}. 
\subsection{From Natural to Art Images}
\label{subsec:ImageNetToArt}
The first feature visualizations we report have been obtained by fine-tuning on the RASTA classification dataset an InceptionV1 architecture pretrained on ImageNet with different sets of hyperparameters. 
\paragraph{Low-level layers are only slightly modified by the fine-tuning.}
The first observation is that low-level layers from the original network trained on ImageNet are hardly modified by the new training on RASTA.  %
This fact will be confirmed by the CKA measure (see \cref{fig:linearCKA_plot}) and the overlapping ratio of the top 100 maximal activation images (see \cref{fig:100_Boxplots_per_layer}) in \cref{sec:CKA_l2_RASTA}. 

\begin{figure}[ht!]
     \begin{center}
    \resizebox{\columnwidth}{!}{
     \begin{tabular}{ cccc  }
 \begin{minipage}[c]{0.24\textwidth} \centering \footnotesize{mixed4c\_3x3\_bottleneck\_pre\_relu:78 ~} \end{minipage}
 & \begin{minipage}[c]{0.24\textwidth} \centering \footnotesize{mixed4d\_pool\_reduce\_pre\_relu:63 ~} \end{minipage}
 &  \begin{minipage}[c]{0.24\textwidth}\centering \footnotesize{mixed4d\_3x3\_pre\_relu:52 ~}   \end{minipage}
  &  \begin{minipage}[c]{0.24\textwidth}\centering \footnotesize{mixed4b\_3x3\_bottleneck\_pre\_relu:35 ~} \end{minipage}\\
  \begin{subfigure}[c]{0.24\textwidth}      \includegraphics[width=\textwidth]{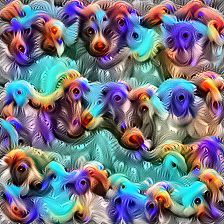} \caption{} \label{fig:MidFilters_Vizu_ImageNet_4c_78} \end{subfigure}
 &\begin{subfigure}[c]{0.24\textwidth}   \includegraphics[width=\textwidth]{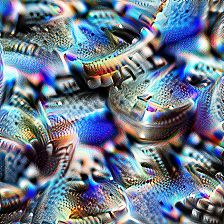}  \caption{} \label{fig:MidFilters_Vizu_ImageNet_4d_63}  \end{subfigure}
 & \begin{subfigure}[c]{0.24\textwidth}  \includegraphics[width=\textwidth]{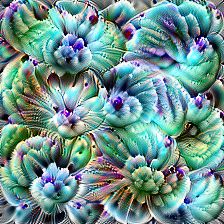} \caption{} \label{fig:MidFilters_Vizu_ImageNet_4d_52}  \end{subfigure}  & \begin{subfigure}[c]{0.24\textwidth}      \includegraphics[width=\textwidth]{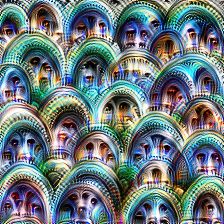} \caption{} \label{fig:MidFilters_Vizu_ImageNet_4b_35}  \end{subfigure}  
 \\
  \begin{subfigure}[c]{0.24\textwidth}      \includegraphics[width=\textwidth]{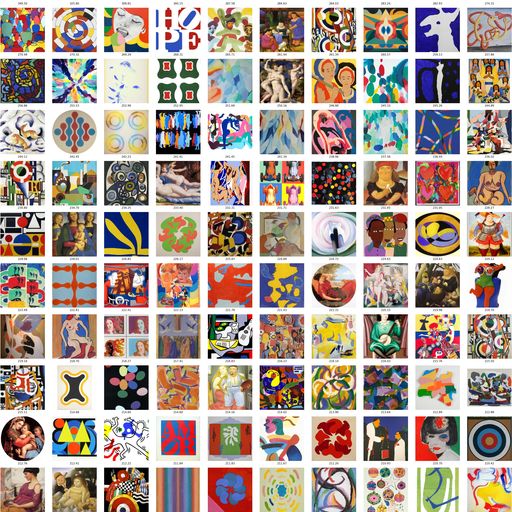} \caption{} \label{fig:AdaptationFiltersRASTA_BlueDrapery_Top100_Init} \end{subfigure}
 &\begin{subfigure}[c]{0.24\textwidth}   \includegraphics[width=\textwidth]{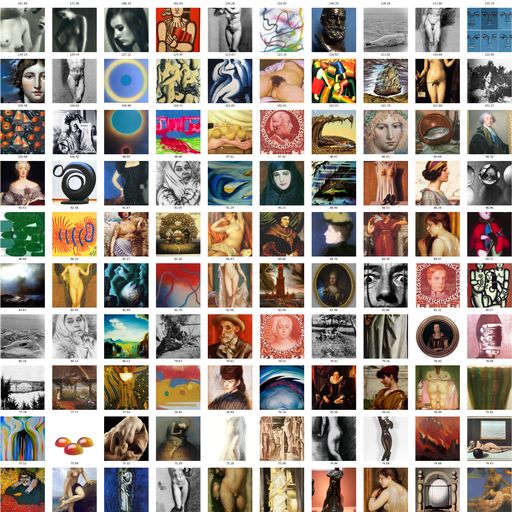} \caption{} \label{fig:AdaptationFiltersRASTA_montagnes_Top100_Init}  \end{subfigure}
 & \begin{subfigure}[c]{0.24\textwidth}  \includegraphics[width=\textwidth]{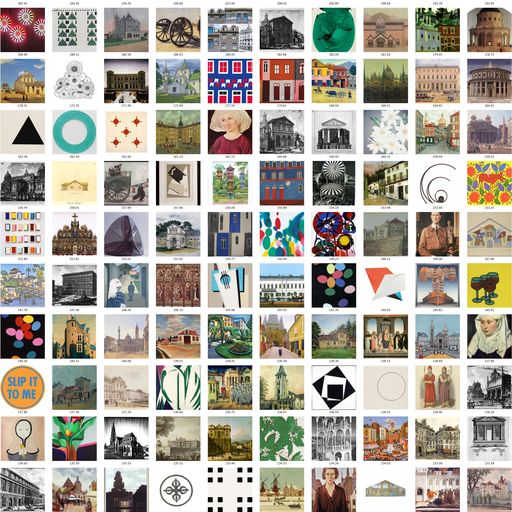} \caption{} \label{fig:AdaptationFiltersRASTA_fronton_Top100_Init}  \end{subfigure} 
 &    \begin{subfigure}[c]{0.24\textwidth}      \includegraphics[width=\textwidth]{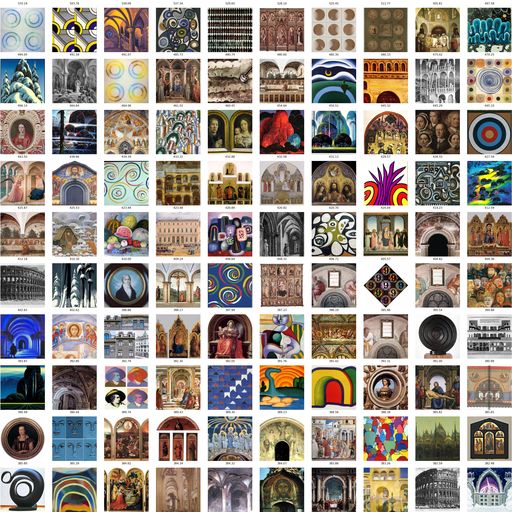} \caption{} \label{fig:AdaptationFiltersRASTA_archs_Top100_Init} \end{subfigure} 
 \\
  \begin{subfigure}[c]{0.24\textwidth}      \includegraphics[width=\textwidth]{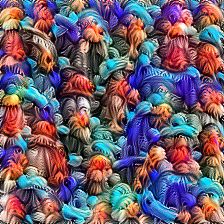} \caption{} 
\label{fig:AdaptationFiltersRASTA_BlueDrapery}
  \end{subfigure}
 &\begin{subfigure}[c]{0.24\textwidth}   \includegraphics[width=\textwidth]{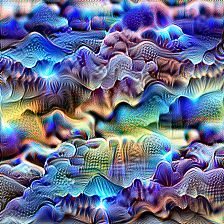}  \caption{} 
 \label{fig:AdaptationFiltersRASTA_montagnes} 
 \end{subfigure}
 & \begin{subfigure}[c]{0.24\textwidth}  \includegraphics[width=\textwidth]{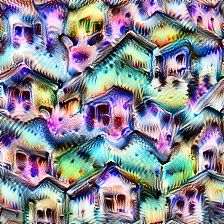} \caption{} 
 \label{fig:AdaptationFiltersRASTA_fronton}
  \end{subfigure} &
    \begin{subfigure}[c]{0.24\textwidth}      \includegraphics[width=\textwidth]{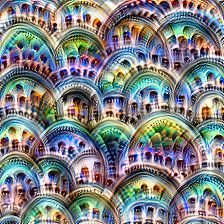} \caption{} \label{fig:AdaptationFiltersRASTA_archs}  \end{subfigure}  \\
  \begin{subfigure}[c]{0.24\textwidth}      \includegraphics[width=\textwidth]{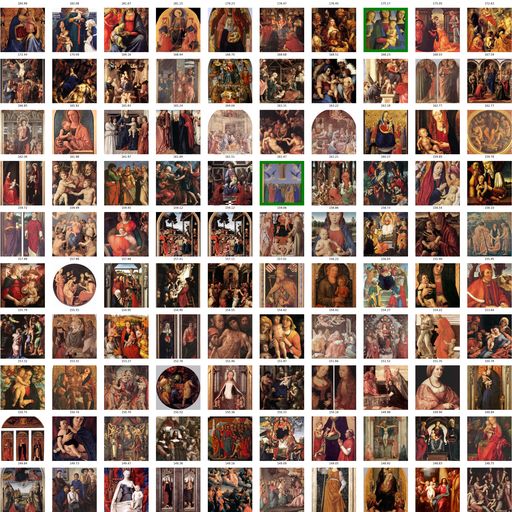} \caption{2\%} \label{fig:AdaptationFiltersRASTA_BlueDrapery_Top100} \end{subfigure}
 &\begin{subfigure}[c]{0.24\textwidth}   \includegraphics[width=\textwidth]{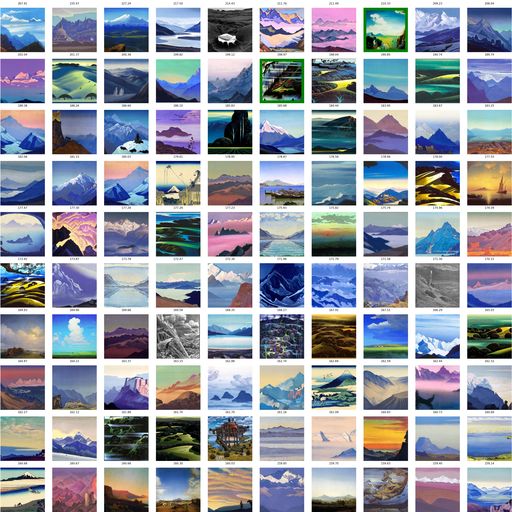} \caption{2\%} \label{fig:AdaptationFiltersRASTA_montagnes_Top100}  \end{subfigure}
 & \begin{subfigure}[c]{0.24\textwidth}  \includegraphics[width=\textwidth]{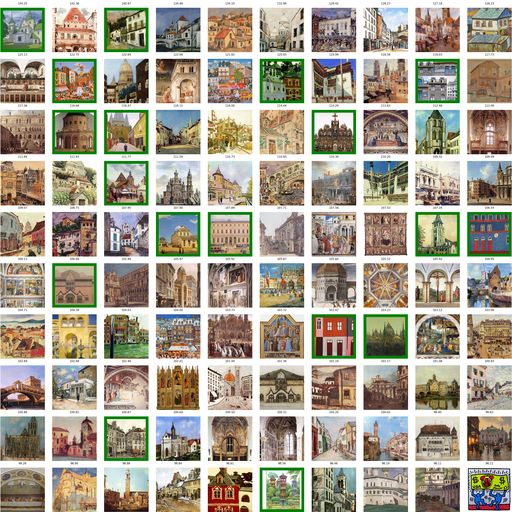} \caption{18\%} \label{fig:AdaptationFiltersRASTA_fronton_Top100}  \end{subfigure} &
 \begin{subfigure}[c]{0.24\textwidth}      \includegraphics[width=\textwidth]{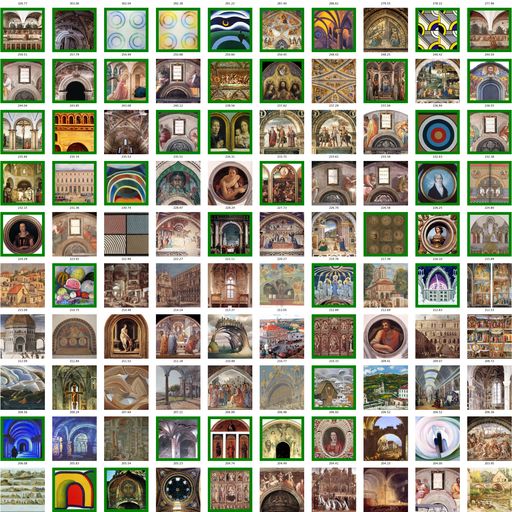} \caption{46\%} \label{fig:AdaptationFiltersRASTA_archs_Top100} \end{subfigure}\\
     \end{tabular}
     }
      \end{center}
\vspace*{-5mm}
\caption{First row: optimized images for InceptionV1 pretrained on ImagneNet.  Second row: top 100 maximal activation examples for the same model. Third and fourth rows: optimized images and maximal activation examples for the same channel of the model fine-tuned on RASTA. The images surrounded by a green line are already present in the top 100 of the pretrained model. The overlapping ratio between the two sets of maximal activation images is displayed at the bottom of each column.}   %
\label{fig:MidFiltersRASTA_Top100Im_And_Vizu}
 \end{figure}

\paragraph{Mid-level layers adapt to the new dataset.} 
Some of the filters have been modified to the specificity of the new dataset by the fine-tuning process, as illustrated in \cref{fig:AdaptationFiltersRASTA_BlueDrapery,fig:AdaptationFiltersRASTA_montagnes,fig:AdaptationFiltersRASTA_fronton}. In these figures are displayed for some channels, the optimized images defined in the second paragraph of \cref{par:featVizu}.
The model learned a red and blue drapery detector, a blue mountain one and a house pediment one. It is worth mentioning that other channels are hardly modified by the fine-tuning process. 
First, among the 70k training samples, some maximal activation images are present in the top 100 both before and after fine-tuning Those images are surrounded by a green line in the last row of \cref{fig:MidFiltersRASTA_Top100Im_And_Vizu}.
Second, in those maximal activation images, we can recognize the pattern that emerged in the optimized image (when we compare the third and last rows). For instance, in the third column of \cref{fig:MidFiltersRASTA_Top100Im_And_Vizu}, a flower-like structure is transformed into a house pediment one. 
Finally, we observe that the detector fine-tuned on RASTA concentrates images with this specific pattern (last row of \cref{fig:MidFiltersRASTA_Top100Im_And_Vizu}). 
The first group of images of the last row contains characters with a blue dress (as the Mary character), the second one blue mountains and the last one buildings depicted with some perspective.
On the other hand, for other channels, the pattern is already present in the optimized image and the detector is slightly adapted to the new dataset. This appears in the form of a minor modification of the optimized image. An arch detector within the pretrained ImageNet model has been modified to detect bigger arches as it can be seen in \cref{fig:AdaptationFiltersRASTA_archs}. The maximal activation images before the fine-tuning already was composed of many buildings images. In this case, the overlapping ratio between the two sets of maximal activation images is equal to 46\%. 
 Nevertheless, we highlight those visualizations but the reader must keep in mind that some channels are not modified by the fine-tuning or are not interpretable at all\footnote{The reader can find more feature visualizations at \url{https://artfinetune.telecom-paris.fr/data/}}.

\begin{figure}[ht!]
    \begin{center}
    \resizebox{\columnwidth}{!}{
    \begin{tabular}{ cccccccccc  }
\multicolumn{10}{c}{Imagenet  Pretrained} \\
\multicolumn{10}{c}{ \includegraphics[width=0.2\textwidth]{im/pretrained/mixed4d_3x3_pre_reluConv2D_52_Imagnet_Deco_toRGB}} \\
\hline
\multicolumn{2}{c}{Mode A} & \multicolumn{2}{c}{Mode B} & \multicolumn{2}{c}{Mode C} & \multicolumn{2}{c}{Mode D}  & \multicolumn{2}{c}{Mode E} \\
\begin{subfigure}[c]{0.2\textwidth}
\caption{Training 1 : 18\%} \label{fig:ModeA_FT1}
\includegraphics[width=\textwidth]{im/RASTA_small01_modif/mixed4d_3x3_pre_reluConv2D_52_RASTA_small01_modif_Deco_toRGB} \end{subfigure}&
\begin{subfigure}[c]{0.2\textwidth}
\caption{Training 2 : 24\%} \label{fig:ModeA_FT2}
\includegraphics[width=\textwidth]{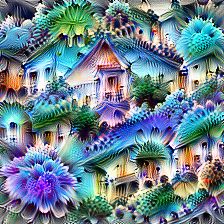} \end{subfigure}&
\begin{subfigure}[c]{0.2\textwidth}
\caption{Training 2 : 34\%} \label{fig:ModeB_FT1}
\includegraphics[width=\textwidth]{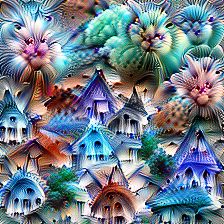} \end{subfigure}&
\begin{subfigure}[c]{0.2\textwidth}
\caption{Training 2 : 42\%} \label{fig:ModeB_FT2}
\includegraphics[width=\textwidth]{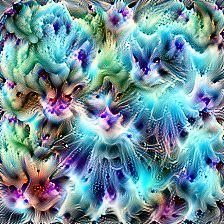} \end{subfigure}&
\begin{subfigure}[c]{0.2\textwidth}
\caption{Training 2 : 22\%} \label{fig:ModeC_FT1}
\includegraphics[width=\textwidth]{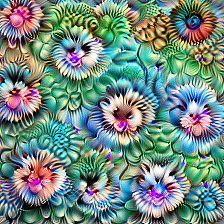} \end{subfigure}&
\begin{subfigure}[c]{0.2\textwidth}
\caption{Training 2 : 8\%} \label{fig:ModeC_FT2}
\includegraphics[width=\textwidth]{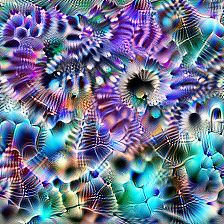}\end{subfigure} &
\begin{subfigure}[c]{0.2\textwidth}
\caption{Training 2 : 10\%} \label{fig:ModeD_FT1}
\includegraphics[width=\textwidth]{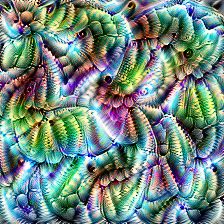} \end{subfigure}&
\begin{subfigure}[c]{0.2\textwidth}
\caption{Training 2 : 13\%} \label{fig:ModeD_FT2}
\includegraphics[width=\textwidth]{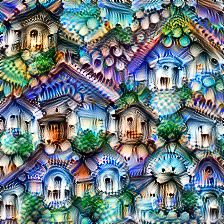} \end{subfigure}&
\begin{subfigure}[c]{0.2\textwidth}
\caption{Training 2 : 2\%} \label{fig:ModeE_FT1}
\includegraphics[width=\textwidth]{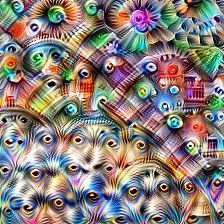} \end{subfigure}&
\begin{subfigure}[c]{0.2\textwidth}
\caption{Training : 3\%} \label{fig:ModeE_FT2}
\includegraphics[width=\textwidth]{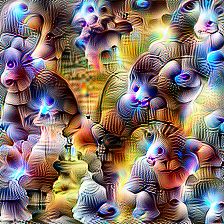} \end{subfigure} \\
    \end{tabular}
    }
     \end{center}
 \vspace*{-3mm}
\caption{Optimized Image for a given channel (mixed4d\_3x3\_pre\_relu:52) with different trainings. The overlapping ratio between the two sets of maximal activation images is displayed on top of the images.}
 \label{tab:SeveralModels_sameFeat_mixed4d_52}
\end{figure}

\paragraph{Learned filters have a high variability.}
We ran 2 distinct fine-tunings for each of the 5 considered optimization schemes named Mode A to E\footnote{See \cref{tab:HyperParams}.}. The initial last layer is different as well as the order of the images in the mini-batches during the training process. 
From a same starting point (the ImageNet weights) but for different hyper-parameters, the training process may sometimes converge to similar optimized images. On the contrary, two optimizations with the same hyper-parameters (same mode) may lead to very different detectors. Those phenomena are illustrated in \cref{tab:SeveralModels_sameFeat_mixed4d_52}.
For this given channel, according to the mode and occurrence of the fine-tuning, one can recognize houses (\cref{fig:ModeA_FT1}), flowers (\cref{fig:ModeC_FT1}), a mix of houses or more abstract patterns (\cref{fig:ModeE_FT1}). 
ImageNet pre-trained filters seem to be a good initialization for learning useful new filters adapted to the artistic style classification and they also permit to learn a variety of new filters.
The percentage of overlap between the set of maximal activation images before and after fine-tuning seems to be correlated to the amount of visual change.%

\begin{figure}[ht!]
     \begin{center}
    \resizebox{\columnwidth}{!}{
     \begin{tabular}{ cccc  }
    \multicolumn{2}{c}{mixed5b\_pool\_reduce\_pre\_relu:92 }  &   \multicolumn{2}{c}{ mixed5b\_5x5\_pre\_relu:82  } \\
       \begin{subfigure}[t]{0.25\textwidth}      \includegraphics[width=\textwidth]{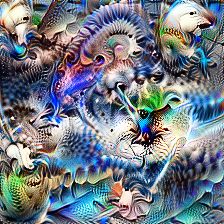} \caption{Optimized Image} \label{fig:HighFilters_Vizu_RASTA_5b_92} \end{subfigure} &
      \begin{subfigure}[t]{0.25\textwidth}         \includegraphics[width=\textwidth]{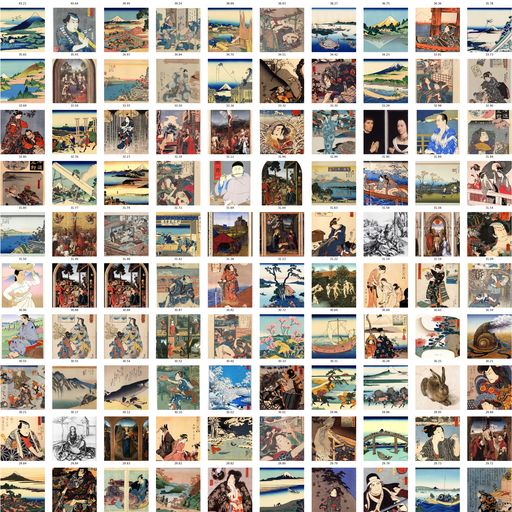} \caption{Maximal activation examples : 1\%} \label{fig:HighLevel_Ukiyoe} \end{subfigure} &  \begin{subfigure}[t]{0.25\textwidth}  \includegraphics[width=\textwidth]{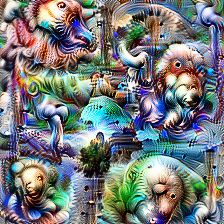} \caption{Optimized Image} \label{fig:HighFilters_Vizu_RASTA_5b_82} \end{subfigure} & 
     \begin{subfigure}[t]{0.25\textwidth}  
    \includegraphics[width=\textwidth]{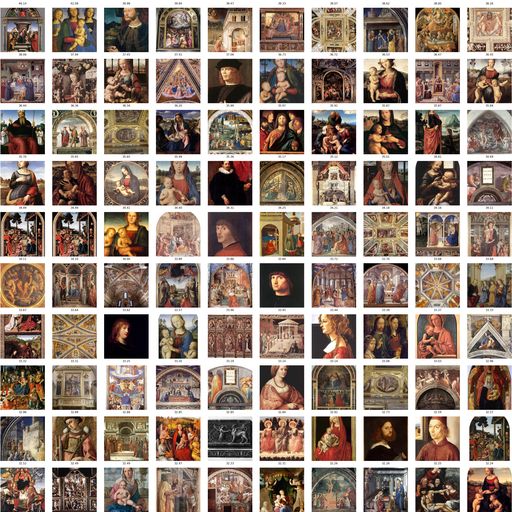} \caption{Maximal activation examples : 0\%} \label{fig:HighLevel_Renaissance}  \end{subfigure}  \\
    \multicolumn{4}{c}{Top 100 composition :} \\
\multicolumn{2}{c}{\small{ Ukiyo-e 82 \% }} &\multicolumn{2}{c}{ \small{Early\_Renaissance 48\%}} \\
\multicolumn{2}{c}{\small{ Northern\_Renaissance 14 \% }} &\multicolumn{2}{c}{ \small{High\_Renaissance 27\%}} \\
\multicolumn{2}{c}{\small{ Early\_Renaissance 3 \% }} &\multicolumn{2}{c}{ \small{Mannerism\_(Late\_Renaissance) 12\%}} \\
     \end{tabular}
    }
      \end{center}
\vspace*{-5mm}
\caption{Optimized Images and Maximal Activation Examples for two high level layers for the model fine-tuned on RASTA. The overlapping ratio between the set of maximal activation images before and after fine-tuning is displayed under the images. The percentage of the 3 most common class is displayed below.}   
\label{fig:HighLevelFiltersRASTA}
 \end{figure}

\paragraph{High-level filters concentrate images from the same classes.} 
The visualizations of high-level layers (near the classification output) are more difficult to interpret, as illustrated in \cref{fig:HighLevelFiltersRASTA}.
The network seems to mix different visual information from the previous layers.
Nevertheless, the group of images with maximal activation for those 2 given channels gather images from the same artistic style after fine-tuning.
The first channel is mostly fired by Ukiyo-e images (\cref{fig:HighLevel_Ukiyoe}), the second one gathers western renaissance artworks (\cref{fig:HighLevel_Renaissance}). There is no visual clue to such clustering in the optimized images. %
In the last image, one may see some green tree in front of a blue sky and some drapery.
The fact that  Early\_Renaissance, High\_Renaissance and Mannerism\_(Late\_Renaissance) classes are clustered together maybe due to their strong visual similarity. Deep model commonly mislabels one of these as another, as mentioned in \cite{lecoutre_recognizing_2017}.
\subsection{Training from Scratch}
\label{subsec:FromScratch}
\begin{figure}[ht!]
     \begin{center}
\resizebox{\columnwidth}{!}{
     \begin{tabular}{ cc|cc  }
 \multicolumn{2}{c|}{The end from scratch} & \multicolumn{2}{c}{All from scratch} \\
 \begin{minipage}[c]{0.24\textwidth} \centering \footnotesize{mixed4d\_5x5\_pre\_relu:50}  \end{minipage}
 & \begin{minipage}[c]{0.24\textwidth} \centering \footnotesize{mixed5a\_3x3\_bottleneck\_pre\_relu:1 ~} \end{minipage}  & \begin{minipage}[c]{0.24\textwidth} \centering \footnotesize{mixed4d:16} \end{minipage} & \begin{minipage}[c]{0.24\textwidth} \centering \footnotesize{mixed4d:66} \end{minipage} \\ 
  \begin{subfigure}[c]{0.24\textwidth}         \includegraphics[width=\textwidth]{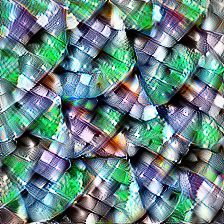}  \caption{}  \label{fig:AdaptationFiltersRASTA_checkboard} \end{subfigure}
 &\begin{subfigure}[c]{0.24\textwidth}         \includegraphics[width=\textwidth]{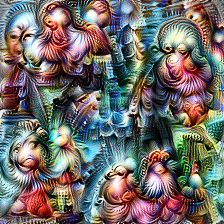}  \caption{}  \label{fig:AdaptationFiltersRASTA_draperie} \end{subfigure} &\begin{subfigure}[c]{0.24\textwidth}   \includegraphics[width=\textwidth]{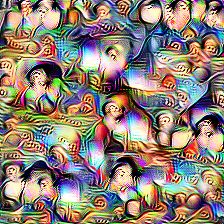}  \caption{} \label{fig:LearnedFromScratch_Ukiyoe}  \end{subfigure}
 & \begin{subfigure}[c]{0.24\textwidth}  \includegraphics[width=\textwidth]{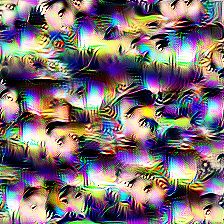} \caption{} \label{fig:LearnedFromScratch_Magic}  \end{subfigure}  \\
 \begin{subfigure}[c]{0.24\textwidth}          \includegraphics[width=\textwidth]{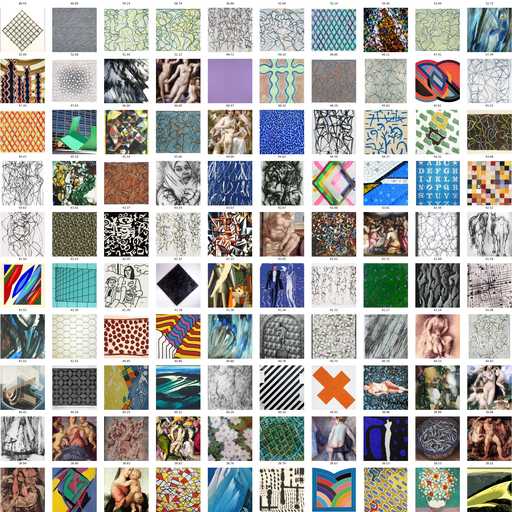}  \caption{ Overlapping : 0\% } \label{fig:RandUnfreeze_Lines}  \end{subfigure}
 &
  \begin{subfigure}[c]{0.24\textwidth}          \includegraphics[width=\textwidth]{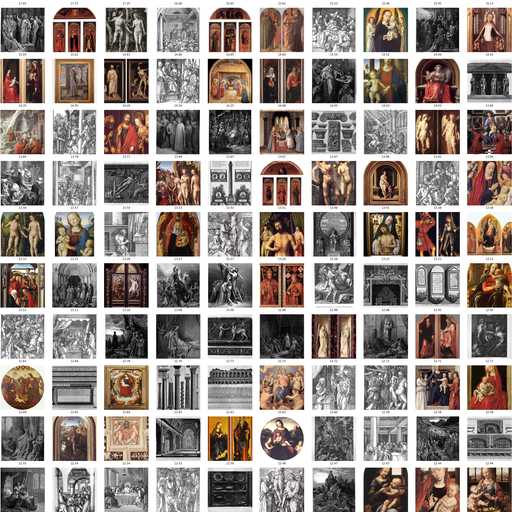}  \caption{0\% } \label{fig:RandUnfreeze_Renaissance}  \end{subfigure}
&
 \begin{subfigure}[c]{0.24\textwidth}          \includegraphics[width=\textwidth]{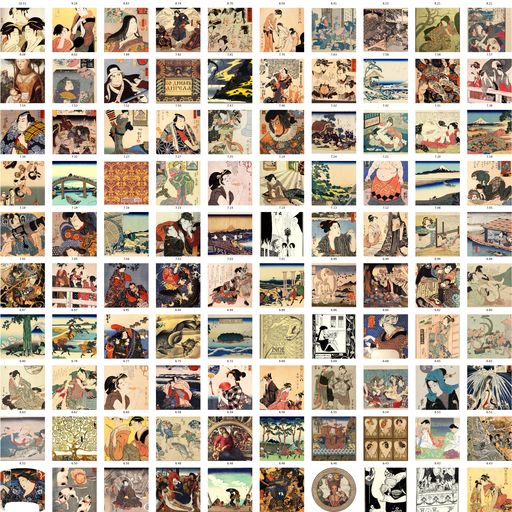}  \caption{0\% } \label{fig:LearnedFromScratch_Top100_Ukiyoe}  \end{subfigure}
  & \begin{subfigure}[c]{0.24\textwidth}  
     \includegraphics[width=\textwidth]{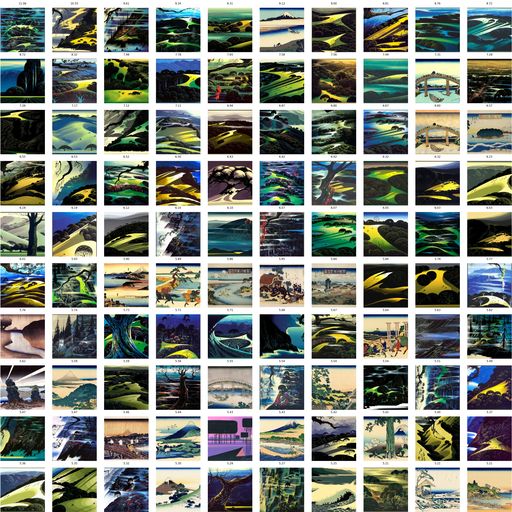}\caption{0\%} \label{fig:LearnedFromScratch_Top100_Magic}  \end{subfigure} \\
     \multicolumn{2}{c|}{Top 100 composition :} &  \multicolumn{2}{c}{Top 100 composition :} \\
     \scriptsize{Abstract\_Expressionism 24\%} &  \scriptsize{  Northern\_Renaissance 39\%}  &          \scriptsize{Ukiyo-e 85\%} &  \scriptsize{ Magic\_Realism 78\% } \\
      \scriptsize{Minimalism 13\%} & \scriptsize{Romanticism 20\%} & \scriptsize{Art\_Nouveau\_(Modern) 11\%} &  \scriptsize{ Ukiyo-e 22\% } \\
      \scriptsize{Art\_Informel 9\%}& \scriptsize{Early_Renaissance 18\%} & \scriptsize{Northern\_Renaissance 2\%} &   \\
     
     \end{tabular}
 }
      \end{center}
\vspace*{-5mm}
\caption{Optimized Image and Maximal activation examples from different mid-level layers. On the left : fine-tuning is performed starting from low-level layers initialized from ImagNet and upper layers initialized at random. On the right, the fine-tuning is fully performed from scratch (randomly initialized layers). The overlapping ratio between the set of maximal activation images before and after fine-tuning is displayed under the images. The percentage of the 3 most common class is displayed below.
}   
\label{fig:RASTA_RandomNets}
 \end{figure}

\paragraph{Mid-level detectors can be learned from scratch when low-level layers are transferred from ImageNet.}
The next experiment consists in fine-tuning a network whose low-level layers are initialized using the training on ImageNet and frozen whereas the mid and high-level layers are initialized randomly. %
In this case, the network is able to learn useful and comprehensible mid-level detectors such as drapery or checkerboard as illustrated in \cref{fig:AdaptationFiltersRASTA_checkboard,fig:AdaptationFiltersRASTA_draperie}.
This phenomenon is most likely triggered by the low-level layers inherited from the ImagNet training, but the emergence of such structured detectors with a relatively small-sized dataset is relatively surprising. 

\paragraph{The optimized images are more difficult to interpret with a full training from scratch.}
A network trained fully from scratch seems yields the same kind of low-level filters that the ones pretrained on ImageNet whereas the mid and high-level layers provide optimized images that are much more difficult to interpret, see \cref{fig:LearnedFromScratch_Ukiyoe,fig:LearnedFromScratch_Magic}.
A possible explanation is that the network may not need to learn very specific filters given its high capacity. 
The training of the network provides filters that are able to fire for a given class such as Ukiyo-e (\cref{fig:LearnedFromScratch_Top100_Ukiyoe}) or Magic\_Realism (\cref{fig:LearnedFromScratch_Top100_Magic}) without being interpretable for humans.

\subsection{Classification Performance.}
\label{sec:perfo_RASTA}
Even though the goal of this work is not to reach the best possible classification performance, we display the corresponding results in \cref{tab:RASTA_accs} to further characterize the considered fine-tuning. From this table, one sees that a simple and short fine-tuning of a pre-trained model yields to a better performance than the off-the-shelf strategy.
The former method is based on extracting features from the ImageNet pretrained model and training the last layer. The features extracted may be too specific to the ImageNet classification task and the classification head too small.  
With training from scratch, we failed to obtain a model as good as the ImageNet pretraining. This can be due to the relatively small size of the RASTA dataset. %
Some data augmentation and a longer training was required to reach 45.29\% Top1 accuracy. 
This experiment confirms Lecoutre et al. \cite{lecoutre_recognizing_2017} conclusions for another deep architecture. 
We conclude this section by observing that a simple way to improve results is to simply  average the prediction of three models trained with different strategies (see last row of  \cref{tab:RASTA_accs}).
\begin{table}[ht!]
 \resizebox{\columnwidth}{!}{
\begin{tabular}{|c|c|c|c|}
\hline 
Method & Top-1 & Top-3 & Top-5 \\ 
\hline 
Off-the-shelf InceptionV1 pretrained on ImageNet  &  30.95 & 58.71 &74.10 \\ %
\hline 
FT of InceptionV1 pretrained on ImageNet (Mode A training 1) & \textbf{55.18} & \textbf{82.25} & \textbf{91.06}\\ %
\hline 
Training from scratch the end of the model with pretrained frozen low-level layers. & 50.35 & 78.04 & 88.42\\ %
\hline 
InceptionV1 trained from scratch & 45.29 & 73.44 & 84.67\\ %
\hline 
Ensemble of the 3 previous models & \textit{58.76} &  \textit{83.99} &  \textit{92.23} \\
\hline
\end{tabular} }
\caption{Top-k accuracies (\%) on RASTA dataset \protect\cite{lecoutre_recognizing_2017} for different methods. The hyperparameters  are different between the methods.}
\label{tab:RASTA_accs}
\end{table}

\subsection{Quantitative Evaluation of the CNNs Modification}
\label{sec:CKA_l2_RASTA}
\begin{figure}[h]
     \begin{center} 
       \resizebox{\textwidth}{!}{
\begin{tikzpicture}

\definecolor{color0}{rgb}{0.215686274509804,0.494117647058824,0.72156862745098}
\definecolor{color1}{rgb}{0.301960784313725,0.686274509803922,0.290196078431373}
\definecolor{color2}{rgb}{0.635294117647059,0.784313725490196,0.925490196078431}
\definecolor{color3}{rgb}{1,0.498039215686275,0}
\definecolor{color4}{rgb}{0.596078431372549,0.305882352941176,0.63921568627451}
\definecolor{color5}{rgb}{0.894117647058824,0.101960784313725,0.109803921568627}
\definecolor{color6}{rgb}{0.968627450980392,0.505882352941176,0.749019607843137}

\begin{axis}[
axis line style={white!80!black},
legend cell align={left},
legend style={fill opacity=0.8, draw opacity=1, text opacity=1, at={(1.01,0.5)}, anchor=west, draw=white!80!black},
tick align=outside,
tick pos=both,
x grid style={white!80!black},
xlabel={Layer},
xmajorgrids,
xmin=-0.55, xmax=11.55,
xtick style={color=white!15!black},
xtick={0,1,2,3,4,5,6,7,8,9,10,11},
xticklabel style = {rotate=45.0},
xticklabels={conv2d0,conv2d1,conv2d2,mixed3a,mixed3b,mixed4a,mixed4b,mixed4c,mixed4d,mixed4e,mixed5a,mixed5b},
y grid style={white!80!black},
ylabel={CKA (linear)},
ymajorgrids,
ymin=0, ymax=1.01,
ytick style={color=white!15!black}
]
\addplot [semithick, color0, dashed, mark=*, mark size=2.5, mark options={solid}]
table {%
0 0.996697902679443
1 0.907883882522583
2 0.904161512851715
3 0.949508428573608
4 0.896285116672516
5 0.854930758476257
6 0.836622774600983
7 0.746934533119202
8 0.623491048812866
9 0.585473418235779
10 0.575650036334991
11 0.373070001602173
};
\addlegendentry{pretrained on ImageNet vs FT on RASTA (Mode A training 1)}
\addplot [semithick, color1, dashed, mark=square*, mark size=2.5, mark options={solid}]
table {%
0 0.999417781829834
1 0.992851853370667
2 0.945651531219482
3 0.974158227443695
4 0.949261605739594
5 0.941104710102081
6 0.909428119659424
7 0.864845156669617
8 0.823457181453705
9 0.779976069927216
10 0.770736575126648
11 0.800441741943359
};
\addlegendentry{FT on RASTA (Mode A training 1) vs FT on RASTA (Mode A training 2)}
\addplot [semithick, color2, dashed, mark=diamond*, mark size=2.5, mark options={solid}]
table {%
0 0.998790681362152
1 0.95068496465683
2 0.958242952823639
3 0.963978707790375
4 0.935011684894562
5 0.903027236461639
6 0.884464144706726
7 0.81660658121109
8 0.761120080947876
9 0.71618515253067
10 0.731996655464172
11 0.741421163082123
};
\addlegendentry{FT on RASTA (Mode A training 1) vs FT on RASTA (Mode B training 1)}
\addplot [semithick, color3, dashed, mark=asterisk, mark size=2.5, mark options={solid}]
table {%
5 0.814802825450897
6 0.656951129436493
7 0.49321237206459
8 0.41984623670578
9 0.347773253917694
10 0.300413757562637
11 0.193351149559021
};
\addlegendentry{The end from scratch vs Random Init}
\addplot [semithick, color4, dashed, mark=triangle*, mark size=2.5, mark options={solid,rotate=180}]
table {%
0 0.729387760162354
1 0.403006553649902
2 0.285842597484589
3 0.405432492494583
4 0.313903570175171
5 0.39902001619339
6 0.407363682985306
7 0.343351125717163
8 0.295531064271927
9 0.249331429600716
10 0.233775988221169
11 0.165569856762886
};
\addlegendentry{From scratch vs Random Init}
\addplot [semithick, color5, dashed, mark=triangle*, mark size=2.5, mark options={solid}]
table {%
5 0.89644193649292
6 0.808457970619202
7 0.723372638225555
8 0.6432244181633
9 0.589040875434875
10 0.524217188358307
11 0.351343214511871
};
\addlegendentry{pretrained on ImageNet vs The end from scratch}
\addplot [semithick, color6, dashed, mark=triangle*, mark size=2.5, mark options={solid,rotate=270}]
table {%
0 0.276158571243286
1 0.556601822376251
2 0.452363759279251
3 0.670311033725739
4 0.610107898712158
5 0.545708477497101
6 0.570127785205841
7 0.549405097961426
8 0.537018418312073
9 0.512823104858398
10 0.496427863836288
11 0.36940661072731
};
\addlegendentry{pretrained on ImageNet vs From scratch}
\end{axis}

\end{tikzpicture}
}
  \end{center}
  \vspace*{-7mm}
    \caption{CKA computed on RASTA test set for different models trained or fine-tuned on RASTA train set.} \label{fig:linearCKA_plot} 
 \end{figure}
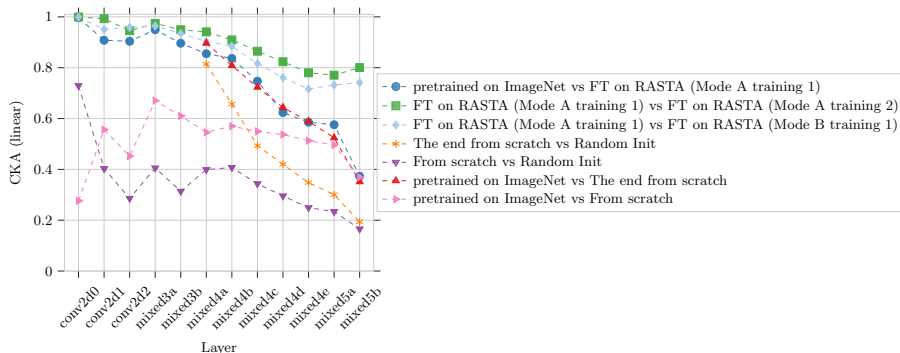

In order to quantify some of the previous observations, we make use of the linear CKA \cite{kornblith_similarity_2019}  as a measure of similarity between two output features at a given layer, for two instances of a network. For computational reason, we used the global spatial average of each channel. The results are shown in \cref{fig:linearCKA_plot}. 
We can observe a decreasing of the CKA when the depth of the layers increases, when we compare the pretrained model with its fine-tuned version (dark blue line). This is a confirmation of what we observed previously with the optimized images (\cref{subsec:ImageNetToArt}).
The fine-tuned models are the closest ones according to the green and light blue lines. The high level layers of those models are closed because those models have been trained on the same dataset from the same initialization point. 
The CKA also decreases with layers when we compare one model from scratch to its random initialization (purple and orange curves).
The values of CKA present here are higher than the ones obtained in \cite{neyshabur_what_2020} for X-ray images. In the case of the model trained from scratch, we even observe several orders of magnitude of difference. This confirms and quantifies the fact that the structure of artistic images is closer to the one of natural images when compared to X-ray images.
 
 \begin{figure}[!tbp]
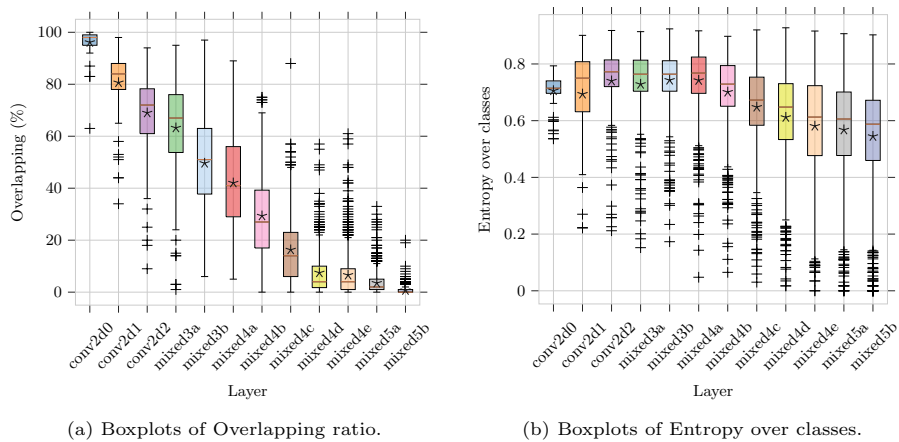

  \centering
  \begin{subfigure}[b]{0.49\textwidth}
  \centering
  \resizebox{\textwidth}{!}{
\input{imInTex/RASTA_small01_modif/Overlapping/OverLap_100_Boxplots_per_layer.tex}
}
    \caption{Boxplots of Overlapping ratio.}
     \label{fig:100_Boxplots_per_layer}
  \end{subfigure}    
      \hfill
     \begin{subfigure}[b]{0.49\textwidth}
  \centering
  \resizebox{\textwidth}{!}{
\input{imInTex/RASTA_small01_modif/Overlapping/Purity_entropy100_Boxplots_per_layer.tex}
}
    \caption{Boxplots of Entropy over classes.} \label{fig:Entropy_class_top100_Boxplots_per_layer}
  \end{subfigure}
 \caption{Boxplots of some metrics on the top 100 maximal activation images for the model fine-tuned on RASTA (Mode A1).
  For each box, the horizontal orange line corresponds to the average result and the star to the median.}
\end{figure}

In addition to feature similarity, we also look at the distance between two models in the parameter space in \cref{tab:RASTA_l2} (as in the recent work \cite{neyshabur_what_2020}). 
We can see that the fine-tuned models are still close one to another and also close to the ImageNet pretrained initialization. In contrast, the models trained from scratch are much farther away from their initialization.
 \begin{table}%
  \resizebox{\columnwidth}{!}{
\begin{tabular}{|c|c|c|}  \hline  
NetA & NetB  &  mean $\ell_{2}$ norm \\ \hline 
Pretrained on ImageNet & FT on RASTA (Mode A training 1) &  1.26\\
FT on RASTA (Mode A training 1) & FT on RASTA (Mode B training 1)&  1.24\\
FT on RASTA (Mode A training 1) & FT on RASTA (Mode A training 2) &  1.23\\
\hline 
The end from scratch & Its Random initialization  &  6.52\\
From scratch & Its Random initialization & 8.13\\
\hline  
 \end{tabular} 
 }
 \caption{Mean over all layers of the $\ell_{2}$ norm of the difference between convolutional kernels between two models} %
\label{tab:RASTA_l2}
\end{table}
We also observe the evolution of the overlapping ratio between the ImageNet pretrained model and the fine-tuned one for the top 100 maximal activation images in \cref{fig:100_Boxplots_per_layer}.  We can see a monotonic decrease of this ratio with the depth of the layer. This is another illustration of the fact that the high level layers are more modified by the fine-tuning. The behavior is the same if we consider the top 1000 maximal activation images. One also observes that channels with low overlapping ratio seem to correspond to optimized images that are more modified by the fine-tuning. This fact should be investigated further and could yield a simple way to browse through optimized images. 
Finally, and in order to quantify the class concentration described in \cref{subsec:ImageNetToArt}, we display the entropy over classes, \cref{fig:Entropy_class_top100_Boxplots_per_layer}), showing a decreasing of the average entropy with the layer depth, starting roughly in the middle of the network architecture.

\subsection{From One Art Dataset to Another}
\label{subsec:ArtToArt}
\Cref{tab:ArtUK_IconArt_perf} compares the classification results obtained for the object classification task on the Paintings dataset  \cite{crowley_search_2014} and on the IconArt dataset \cite{gonthier_weakly_2018} when using a model pretrained on ImageNet or fine-tuned on RASTA dataset (from an ImageNet initialization or from scratch). On the contrary to \cite{sabatelli_deep_2018} where they compared the pretraining on the Rijkmuseum dataset and ImageNet for the Antwerp dataset, the task is not the same between the two artistic datasets: classification of the artistic style versus object classification. 
Once again the fine-tuning strategy is better than the off-the-shelf one.  
The most important observation is that the double fine-tuning (first using RASTA, then using the considered dataset) outperforms the direct fine-tuning using only the dataset at hand. 
The filters learned on RASTA seem to be more adapted to other artistic datasets and ease the transfer in these two cases (IconArt and Paintings) where the datasets are relatively small. 
Finally, a model only trained on RASTA (last row of the two tables) will not provide a good initialization point for fine-tuning, neither for IconArt, nor for Paintings. This is most probably due to the size of the RASTA dataset. 
\begin{table}
 \resizebox{\columnwidth}{!}{
\begin{tabular}{|c|c|c|}
\hline
 Method & Paintings &  IconArt \\
 \hline
Off-the-shelf InceptionV1 pretrained on ImageNet  & 56.0  & 53.2  \\ 
Off-the-shelf InceptionV1 pretrained on ImageNet and RASTA & 52.4  & 54.4  \\ 
\hline 
FT of InceptionV1 pretrained on ImageNet (Mode A training 1) &  64.8 & 59.2 \\
\hline
FT of InceptionV1 pretrained on ImageNet and RASTA & \textbf{65.6} & \textbf{67.4}  \\
FT of InceptionV1 trained from scratch on RASTA the end of the model with pretrained frozen low-level &  59.6 & 59.4  \\
FT of InceptionV1 trained from scratch on RASTA  & 49.1 &  50.1 \\
\hline
\end{tabular} 
}
\caption{Mean Average Precision (\%) on Paintings  \protect\cite{crowley_search_2014} and IconArt test sets  \protect\cite{gonthier_weakly_2018}.}
\label{tab:ArtUK_IconArt_perf}
\end{table}

In \cref{tab:SmallArtDataset_cka_l2}, we use the two previously mentioned metrics to compare the different models fine-tuned on the IconArt and Paintings datasets.
The model fine-tuned on a small art dataset will stay similar to its ImageNet pretrained initialization (with a CKA of 0.89 or 0.91 for the IconArt and Paintings datasets). A fine-tuning on the large RASTA dataset changes more the network ($CKA=0.77$ and $\ell_2$ norm = 1.26). A double fine-tuning permits to go even further from the original pretrained weights ($CKA=0.73$ and $0.76$). As already mentioned, this method provides the best classification performance. 
 
In the case of the model trained from scratch (two last lines of \cref{tab:SmallArtDataset_cka_l2}), the change between initialization and optimal model is also large due to the randomness of the initialization but those models are worst in terms of classification. 

\begin{savenotes}
\begin{table}[ht!]
  \resizebox{\columnwidth}{!}{
\begin{tabular}{|c|c|cc|cc|}  \hline  
\multicolumn{2}{|c|}{Nets / Small art dataset used :} & \multicolumn{2}{c|}{IconArt} & \multicolumn{2}{c|}{Paintings} \\   \hline
NetA & NetB  &   mean CKA & mean $\ell_2$ norm &   mean CKA & mean $\ell_2$ norm \\ \hline  
Pretrained on ImageNet & FT on small art dataset & 0.90 & 0.14 & 0.91 & 0.15  \\
Pretrained on ImageNet & FT on RASTA + FT on small dataset & 0.73 & 1.61 & 0.76 & 1.67  \\
 FT on RASTA (Mode A) & FT on RASTA + FT on small dataset  & 0.79 & 0.78  & 0.77 & 0.86\\
 \hline 
 The end from scratch on RASTA &The end from scratch on RASTA + FT on small dataset &  0.70 & 0.91 &  0.72  &  1.01 \\
  From scratch on RASTA & From scratch on RASTA + FT on small dataset &  0.83 & 0.27 &  0.79 & 0.52 \\
\hline  
 \end{tabular} 
}
 \caption{Mean linear CKA (on IconArt or Paintings test set) and mean $\ell_2$ norm between models.}
\label{tab:SmallArtDataset_cka_l2}
\end{table}
\end{savenotes}

\vspace*{-5mm}

\section{Conclusion}

In this work, we have investigated the effect of fine-tuning a network pre-trained on ImageNet using artistic datasets. We made use of visualization techniques and quantitative assessments of the changes of the networks. Among other things, we have shown that some of the intermediate layers of the networks exhibit easily recognizable patterns that appear to be more related to art images than the patterns learned on natural images, while lower layers of the network are hardly changed. We have also shown that higher layers tend to concentrate classes after fine-tuning. Eventually, we have also shown that a double fine-tuning involving a medium size artistic dataset can help the classification of small-sized artistic datasets in accordance with visual patterns more related to the domain. The classification tasks between the two artistic datasets do not need to be identical. In our case, the intermediate task is style classification whereas the final one is object classification. This study provides some insights on the way networks are modified by fine-tuning in the case of artistic databases. Several of the findings in this work necessitate further confirmation, possible on larger databases \cite{strezoski_omniart_2018,wilber_bam_2017}.  Another perspective would be to go further in the use of feature visualization, as it is done in \cite{olah_overview_2020,olah_building_2018}.
For instance, it could be more informative to look at the patches that fire a given channel rather than the whole input image as in \cite{olah_building_2018}. In \cite{olah_overview_2020}, the authors claim for universality in different deep convolutional architectures and it is of interest to check if the same is true for artistic datasets.

\bibliographystyle{splncs04}
\bibliography{references}

 \clearpage
\begin{subappendices}
\renewcommand{\thesection}{\Alph{section}}%
 \appendix
\section{Experiment Setup} 
 
 The hyperparameters of the different training schemes can be found in \cref{tab:HyperParams}. The hyperparameters we have varied are the learning rate for the dense classification layer, the learning rate for the convolutional layers, the use of a deep supervision (adding auxiliary supervision branches after certain intermediate layers during training), the number of unfrozen layers (i.e. trainable layers), the maximum number of epochs, the optimizer. We also consider two different data augmentation scheme, one named "Small transformation" which consists in small geometric transformation (horizontal flip and/or translation of 28 pixels maximum) and one named "Random crops" which consists in taking random crop of size 224$\times$224 within an image resized to 256 for its smallest dimension. 
 Finally, we also may have considered a learning rate schedule similar to the one used in the original paper of the InceptionV1 model  \cite{szegedy_going_2015}.
\begin{table}[ht!]
   \resizebox{\columnwidth}{!}{
 \begin{tabular}{|c|c|c|c|c|c|c|c|c|}
 \hline 
 Mode & Learning rate & Learning rate & Deep Supervision & Maximum number & Number of unfrozen & Data  & Optimizer & Learning rate \\ 
 & on the last dense layer & on the other layers & & of epochs & layers & Augmentation & & schedule \\
 \hline 
 \hline 
 A & 0.01 & 0.001 & No & 20 & All & No & SGD & No \\ 
 \hline 
 B & 0.001 & 0.0001 & No  & 20 & All & No & SGD & No \\ 
 \hline 
 C & 0.001 & 0.001 & No  & 20 & All & No & SGD & No \\ 
 \hline 
 D & 0.001 & 0.0001 & Yes & 20 & All & No & SGD & No \\ 
 \hline 
 E & 0.001 & 0.001 & Yes  & 20 & All & No & SGD & No \\ 
 \hline
  F  & 0.01 & 0.01 & No & 20 & All & No & SGD & No \\ 
 \hline \hline  
 High layers from scratch & 0.0001 &  0.0001 & No & 200 & 50 & Small transformation & Adam  & No \\ 
 \hline  \hline 
 From scratch & 0.001 & 0.001 & Yes  & 200 & All & Random crops & SGD  & As in \cite{szegedy_going_2015} \\ 
 \hline 
 \end{tabular}
 }
 \caption{Hyperparameters of the different training schemes.}
 \label{tab:HyperParams}
 \end{table}

\section{Additional Figures}

\subsection{From Natural to Art Images}
\begin{figure}[ht!]
      \begin{center}
      \resizebox{\columnwidth}{!}{
      \begin{tabular}{ cccc  }
 \begin{minipage}[c]{0.24\textwidth} \centering Channel Name :   \end{minipage}  
  & \begin{minipage}[c]{0.24\textwidth} \centering conv2d1  \_pre\_relu:30 ~  \end{minipage}
  & \begin{minipage}[c]{0.24\textwidth} \centering mixed3a\_3x3\_pre\_relu:12~  \end{minipage}
  &  \begin{minipage}[c]{0.24\textwidth}\centering mixed3a\_5x5\_bottleneck\_pre\_relu:8~   \end{minipage} \\
 \begin{minipage}[c]{0.24\textwidth} \centering  Imagenet  Pretrained  Optimized Image  \end{minipage}  &
   \begin{subfigure}[c]{0.24\textwidth}      \includegraphics[width=\textwidth]{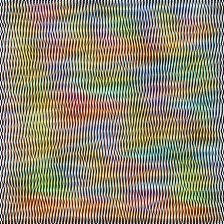} \caption{} \label{fig:LowFilters_ImageNet_conv2d_30} \end{subfigure}
  &\begin{subfigure}[c]{0.24\textwidth}     \includegraphics[width=\textwidth]{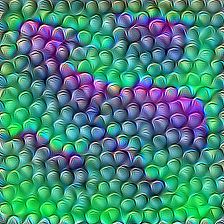}\caption{} \label{fig:LowFilters_ImageNet_3a_12}  \end{subfigure}
  & \begin{subfigure}[c]{0.24\textwidth}  \includegraphics[width=\textwidth]{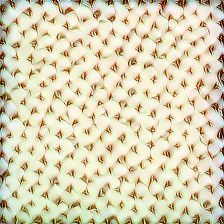}\caption{} \label{fig:LowFilters_ImageNet_3a_8}  \end{subfigure} \\
  \begin{minipage}[c]{0.24\textwidth} \centering   RASTA Fine Tuned  Optimized Image  \end{minipage}  &
   \begin{subfigure}[c]{0.24\textwidth}  
     \includegraphics[width=\textwidth]{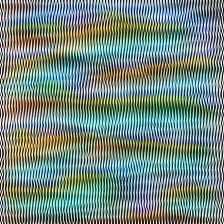} \caption{} \label{fig:LowFilters_RASTA_conv2d_30} \end{subfigure}
  &\begin{subfigure}[c]{0.24\textwidth}     \includegraphics[width=\textwidth]{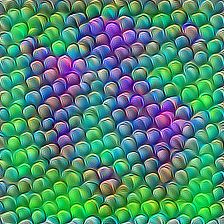} \caption{} \label{fig:LowFilters_RASTA_3a_12}  \end{subfigure}
  & \begin{subfigure}[c]{0.24\textwidth}  
     \includegraphics[width=\textwidth]{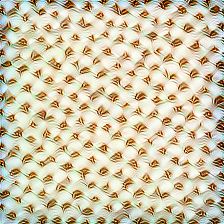} \caption{} \label{fig:LowFilters_RASTA_3a_8}  \end{subfigure} \\
      \end{tabular}
     }
       \end{center}
 \caption{Optimized Images for one individual channel from different low-level layers. First row InceptionV1 pretrained on ImagneNet, second row fine-tuned on RASTA.}   
 \label{fig:lowLevelFiltersRASTA}
  \end{figure}

\begin{figure}[ht!]
     \begin{center}
     \begin{tabular}{ ccc  }
\begin{minipage}[c]{0.28\textwidth} \centering Channel Name :   \end{minipage}  
 & \begin{minipage}[c]{0.28\textwidth} \centering mixed4b:361   \end{minipage}
 &  \begin{minipage}[c]{0.28\textwidth}\centering mixed4d:106   \end{minipage} \\
\begin{minipage}[c]{0.28\textwidth} \centering  Imagenet  Pretrained  Optimized Image  \end{minipage}  
&  \begin{subfigure}[c]{0.28\textwidth}      \includegraphics[width=\textwidth]{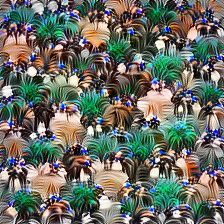} \caption{} \label{fig:MidFilters_Vizu_ImageNet_4b_361} \end{subfigure}
 & \begin{subfigure}[c]{0.28\textwidth}  \includegraphics[width=\textwidth]{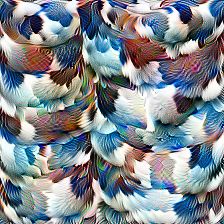} \caption{} \label{fig:MidFilters_Vizu_ImageNet_4d_106}  \end{subfigure} \\
 \begin{minipage}[c]{0.28\textwidth} \centering  Imagenet  Pretrained  Maximal Activation Examples  \end{minipage}  &
  \begin{subfigure}[c]{0.28\textwidth}      \includegraphics[width=\textwidth]{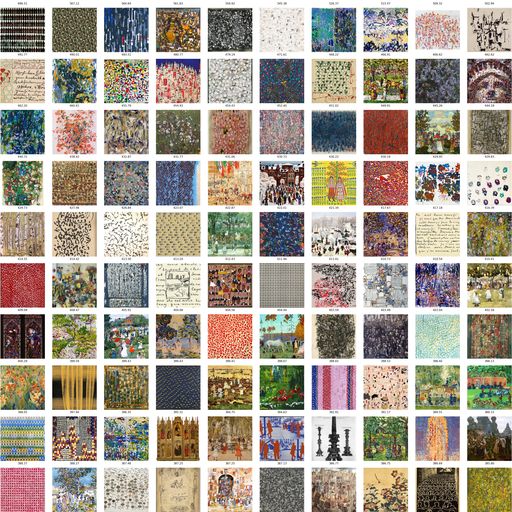} \caption{} \label{fig:MidFilters_Top100_ImageNet_4b_361} \end{subfigure}
 & \begin{subfigure}[c]{0.28\textwidth}  \includegraphics[width=\textwidth]{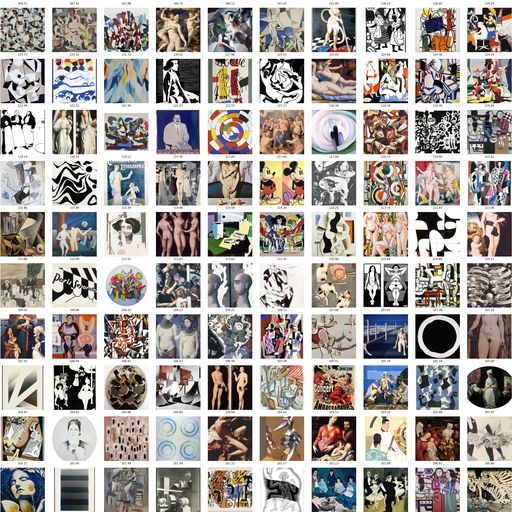} \caption{} \label{fig:MidFilters_Top100_ImageNet_4d_106}  \end{subfigure} \\
\begin{minipage}[c]{0.28\textwidth} \centering   RASTA Fine Tuned Optimized Image\end{minipage} 
&  \begin{subfigure}[c]{0.28\textwidth}      \includegraphics[width=\textwidth]{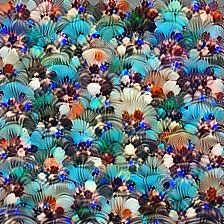} \caption{} \label{fig:MidFilters_Vizu_RASTA_4b_361} \end{subfigure}
 & \begin{subfigure}[c]{0.28\textwidth}  \includegraphics[width=\textwidth]{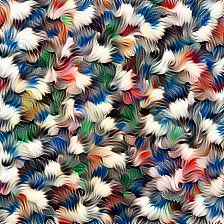} \caption{} \label{fig:MidFilters_Vizu_RASTA_4d_106}  \end{subfigure} \\
 \begin{minipage}[c]{0.28\textwidth} \centering  RASTA  Fine Tuned Maximal Activation Examples \end{minipage}   &
  \begin{subfigure}[c]{0.28\textwidth}      \includegraphics[width=\textwidth]{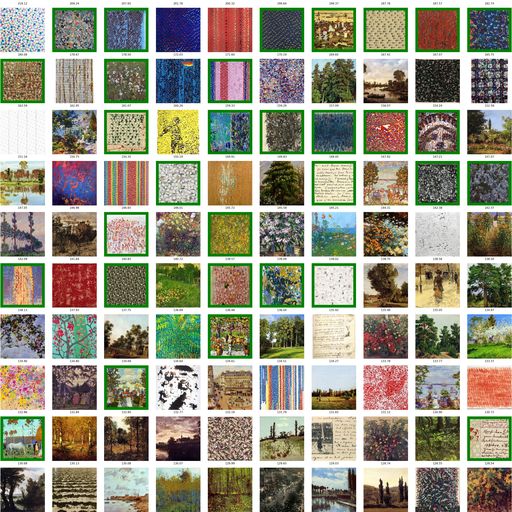} \caption{} \label{fig:MidFilters_Top100_RASTA_4b_361} \end{subfigure}
 & \begin{subfigure}[c]{0.28\textwidth}  \includegraphics[width=\textwidth]{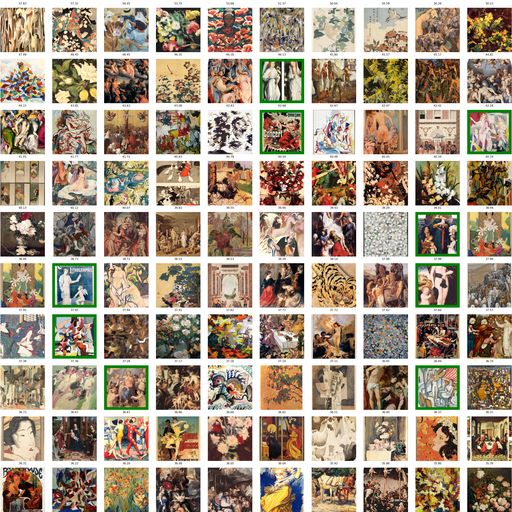} \caption{} \label{fig:MidFilters_Top100_RASTA_4d_106}  \end{subfigure} \\
  Overlapping & 31\%  & 9\% \\
     \end{tabular}
      \end{center}
\caption{Optimized Image and Maximal Activation Examples for one Individual channel. First and second rows InceptionV1 pre-trained on ImagneNet, third and fourth rows fine-tuned on RASTA. The images surrounded by a green line are already presented in the top 100 of the pre-trained model. The percentage of images in common between the two sets of maximal activation images is displayed at the bottom of each column.}    %
\label{fig:MidFiltersRASTA_Top100Im_And_Vizu_2}
 \end{figure}

\paragraph{Low-level layers.}
The low-level layers from the original network trained on ImageNet are hardly modified by the new training on RASTA \cite{lecoutre_recognizing_2017} (see \cref{fig:lowLevelFiltersRASTA}). 
Note that in \cite{yin_object_2016}, it has been shown that the low-level layer filters have been modified by a fine-tuning on an almost monochrome drawing training dataset. This suggests that the statistics of painting images are closer to those of natural images than those of drawing ones.
\paragraph{Mid-level layers.} 
Two other mid-level channels are shown in \cref{fig:MidFiltersRASTA_Top100Im_And_Vizu_2}.
 
\begin{figure}[ht!]
     \begin{center}
     \begin{tabular}{ cc }
\multicolumn{2}{c}{Imagenet  Pretrained} \\
\multicolumn{2}{c}{ \includegraphics[width=0.2\textwidth]{im/pretrained/mixed4d_3x3_pre_reluConv2D_52_Imagnet_Deco_toRGB}  \includegraphics[width=0.2\textwidth]{im/pretrained/ActivationsImages/RASTA_mixed4d_3x3_pre_relu_52_Most_Pos_Images_NumberIm100_meanAfterRelu} } \\
\hline
\begin{subfigure}[c]{0.41\textwidth}
\caption{Mode A training 1,  Overlapping : 18\%} \label{fig:ModeA_FT1_A}  
\vspace*{-2mm}
\includegraphics[width=0.49\textwidth]{im/RASTA_small01_modif/mixed4d_3x3_pre_reluConv2D_52_RASTA_small01_modif_Deco_toRGB} 
\includegraphics[width=0.49\textwidth]{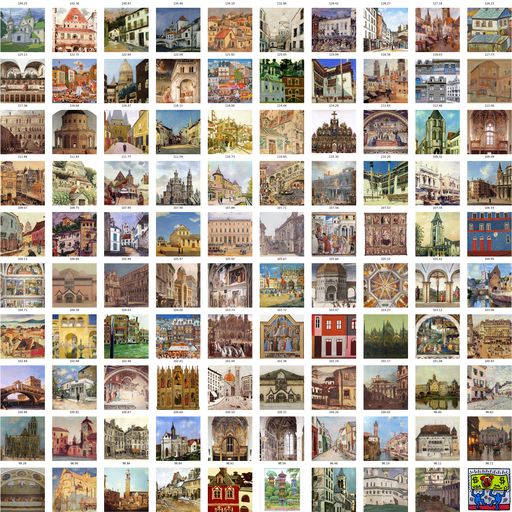}
\end{subfigure} & 
\begin{subfigure}[c]{0.41\textwidth}
\caption{Mode A training 2, Overlapping : 24\%} \label{fig:ModeA_FT2_A}  
\vspace*{-2mm}
\includegraphics[width=0.49\textwidth]{im/RASTA_small01_modif1/mixed4d_3x3_pre_reluConv2D_52_RASTA_small01_modif1_Deco_toRGB}
\includegraphics[width=0.49\textwidth]{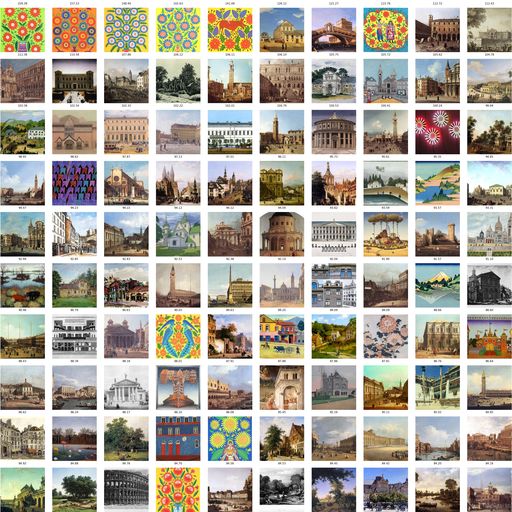}
\end{subfigure} \\
\begin{subfigure}[c]{0.41\textwidth}
\caption{Mode B training 1, Overlapping : 34\%} \label{fig:ModeB_FT1_A}  
\vspace*{-2mm}
\includegraphics[width=0.49\textwidth]{im/RASTA_small001_modif/mixed4d_3x3_pre_reluConv2D_52_RASTA_small001_modif_Deco_toRGB}
\includegraphics[width=0.49\textwidth]{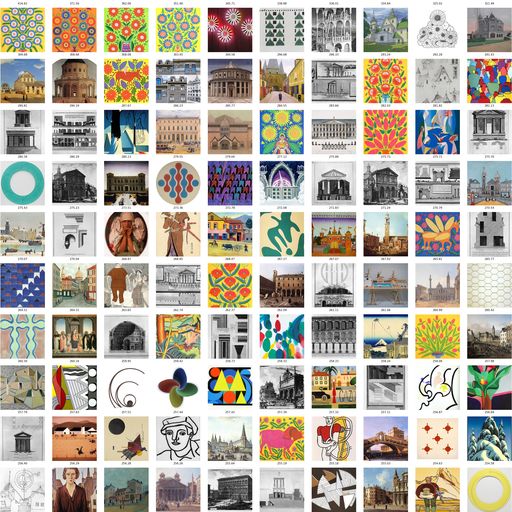}
\end{subfigure} & 
\begin{subfigure}[c]{0.41\textwidth}
\caption{Mode B training 2, Overlapping : 42\%} \label{fig:ModeB_FT2_A}  
\vspace*{-2mm}
\includegraphics[width=0.49\textwidth]{im/RASTA_small001_modif1/mixed4d_3x3_pre_reluConv2D_52_RASTA_small001_modif1_Deco_toRGB}
\includegraphics[width=0.49\textwidth]{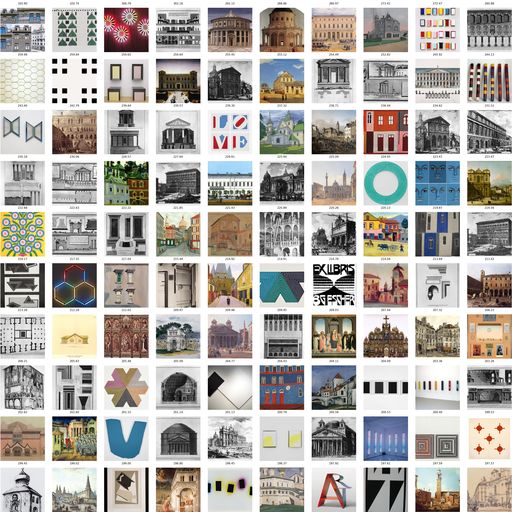}
\end{subfigure} \\
\begin{subfigure}[c]{0.41\textwidth}
\caption{Mode C training 1, Overlapping : 22\%} \label{fig:ModeC_FT1_A}  
\vspace*{-2mm}
\includegraphics[width=0.49\textwidth]{im/RASTA_big001_modif/mixed4d_3x3_pre_reluConv2D_52_RASTA_big001_modif_Deco_toRGB}
\includegraphics[width=0.49\textwidth]{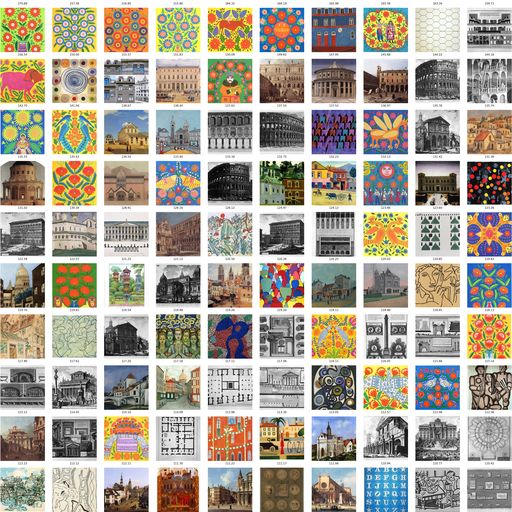}
\end{subfigure} & 
\begin{subfigure}[c]{0.41\textwidth}
\caption{Mode C training 2, Overlapping : 8\%} \label{fig:ModeC_FT2_A}  
\vspace*{-2mm}
\includegraphics[width=0.49\textwidth]{im/RASTA_big001_modif1/mixed4d_3x3_pre_reluConv2D_52_RASTA_big001_modif1_Deco_toRGB}
\includegraphics[width=0.49\textwidth]{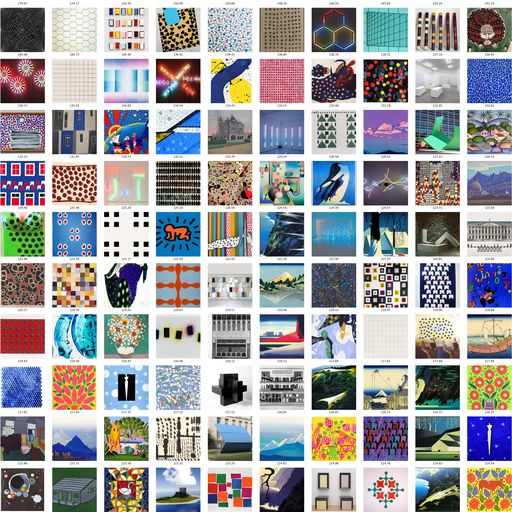}
\end{subfigure} \\
\begin{subfigure}[c]{0.41\textwidth}
\caption{Mode D training 1, Overlapping : 10\%} \label{fig:ModeD_FT1_A}  
\vspace*{-2mm}
\includegraphics[width=0.49\textwidth]{im/RASTA_small001_modif_deepSupervision/mixed4d_3x3_pre_reluConv2D_52_RASTA_small001_modif_deepSupervision_Deco_toRGB} 
\includegraphics[width=0.49\textwidth]{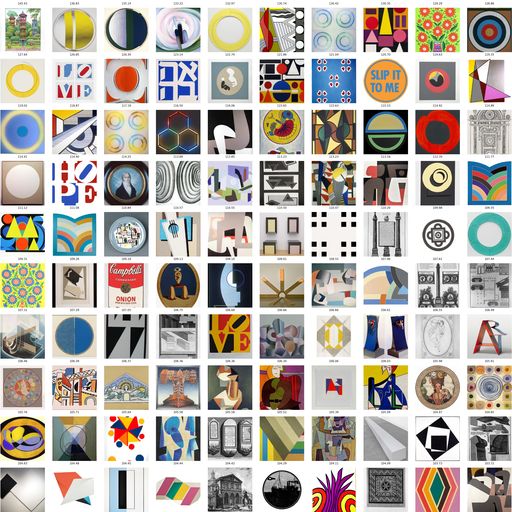} 
\end{subfigure} & 
\begin{subfigure}[c]{0.41\textwidth}
\caption{Mode D training 2, Overlapping : 13\%} \label{fig:ModeD_FT2_A}  
\vspace*{-2mm}
\includegraphics[width=0.49\textwidth]{im/RASTA_small001_modif_deepSupervision1/mixed4d_3x3_pre_reluConv2D_52_RASTA_small001_modif_deepSupervision1_Deco_toRGB} 
\includegraphics[width=0.49\textwidth]{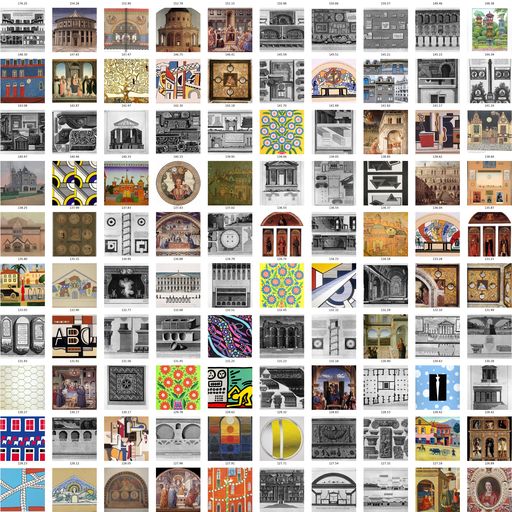} 
\end{subfigure} \\
\begin{subfigure}[c]{0.41\textwidth}
\caption{Mode E training 1, Overlapping : 2\%} \label{fig:ModeE_FT1_A}  
\vspace*{-2mm}
\includegraphics[width=0.49\textwidth]{im/RASTA_big001_modif_deepSupervision/mixed4d_3x3_pre_reluConv2D_52_RASTA_big001_modif_deepSupervision_Deco_toRGB}  
\includegraphics[width=0.49\textwidth]{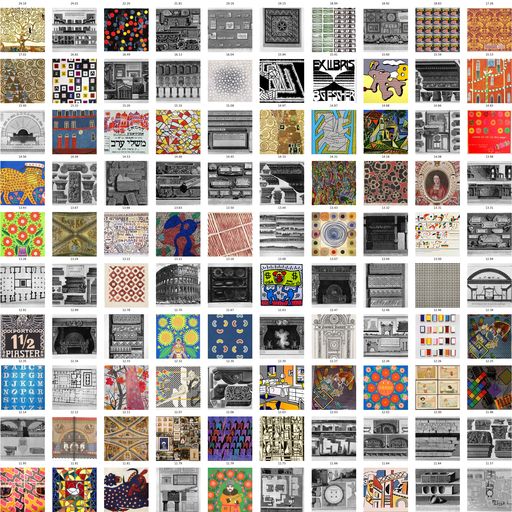}
\end{subfigure} & 
\begin{subfigure}[c]{0.41\textwidth}
\caption{Mode E training 2, Overlapping : 3\%} \label{fig:ModeE_FT2_A} 
\vspace*{-2mm} 
\includegraphics[width=0.49\textwidth]{im/RASTA_big001_modif_deepSupervision1/mixed4d_3x3_pre_reluConv2D_52_RASTA_big001_modif_deepSupervision1_Deco_toRGB} 
\includegraphics[width=0.49\textwidth]{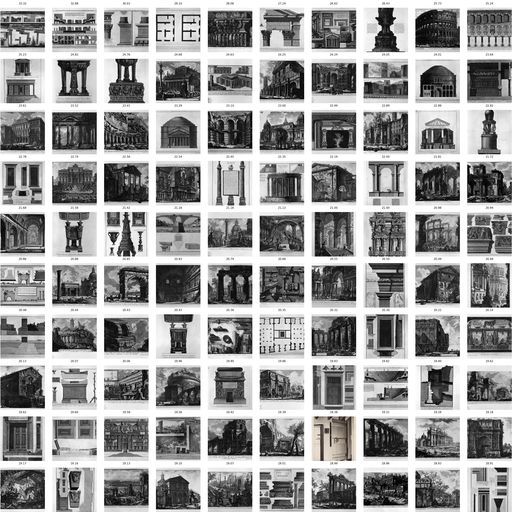} 
\end{subfigure} 
     \end{tabular}
      \end{center}
  \vspace*{-3mm}
\caption{Optimized Image and Maximal Activation Examples for a given channel (mixed4d\_3x3\_pre\_relu:52) with different training. The overlapping between the two sets of maximal activation images is displayed on top of each pair of images.}
 \label{tab:SeveralModels_sameFeat_mixed4d_52_appendix}
 \end{figure}

The top 100 maximal activation images associated with the optimized images shown in \cref{tab:SeveralModels_sameFeat_mixed4d_52} can be found in \cref{tab:SeveralModels_sameFeat_mixed4d_52_appendix}
 The set of maximal activation images concentrates images with a similar pattern (second and fourth column of  \cref{tab:SeveralModels_sameFeat_mixed4d_52_appendix}).
The new filters obtained with Mode A Fine-tuning 2 or Mode B Fine-tuning 1 are fired by colorful flowers and building images (\cref{fig:ModeA_FT2_A,fig:ModeB_FT1_A}).
The one obtained with Mode E Fine-Tuning 2 (\cref{fig:ModeE_FT2_A}) only detects black and white images and the one from Mode A Fine-tuning 1 is only fired by building images  (\cref{fig:ModeA_FT1_A}).

\begin{figure}[ht!]
     \begin{center}
     \resizebox{\columnwidth}{!}{
     \begin{tabular}{ cccc  }
\begin{minipage}[c]{0.24\textwidth} \centering Channel Name :   \end{minipage}  
 & \begin{minipage}[c]{0.24\textwidth} \centering mixed5b\_pool  \_reduce\_pre\_relu:92   \end{minipage}
 & \begin{minipage}[c]{0.24\textwidth} \centering mixed5b\_3x3\_pre\_relu:33  \end{minipage}
 &  \begin{minipage}[c]{0.24\textwidth}\centering mixed5b\_5x5\_pre\_relu:82   \end{minipage} \\
\begin{minipage}[c]{0.24\textwidth} \centering  Imagenet  Pretrained Optimized Image \end{minipage}  &
  \begin{subfigure}[c]{0.24\textwidth}      \includegraphics[width=\textwidth]{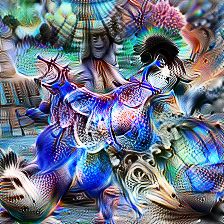} \caption{} \label{fig:HighFilters_Vizu_ImageNet_5b_92_Appendix} \end{subfigure}
 &\begin{subfigure}[c]{0.24\textwidth}   \includegraphics[width=\textwidth]{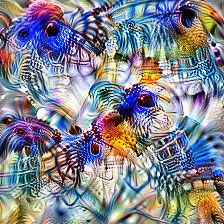}  \caption{} \label{fig:HighFilters_Vizu_ImageNet_5b_33_Appendix}  \end{subfigure}
 & \begin{subfigure}[c]{0.24\textwidth}  \includegraphics[width=\textwidth]{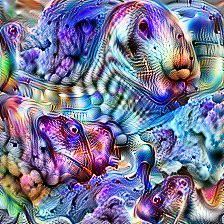} \caption{} \label{fig:HighFilters_Vizu_ImageNet_5b_82_Appendix}  \end{subfigure} \\
 \begin{minipage}[c]{0.24\textwidth} \centering   RASTA Fine Tuned Optimized Image \end{minipage}  &
  \begin{subfigure}[c]{0.24\textwidth}      \includegraphics[width=\textwidth]{im/RASTA_small01_modif/mixed5b_pool_reduce_pre_reluConv2D_92_RASTA_small01_modif_Deco_toRGB} \caption{} \label{fig:HighFilters_Vizu_RASTA_5b_92_Appendix} \end{subfigure}
 &\begin{subfigure}[c]{0.24\textwidth}   \includegraphics[width=\textwidth]{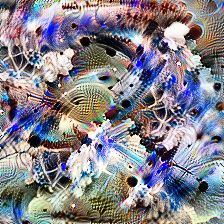}  \caption{} \label{fig:HighFilters_Vizu_RASTA_5b_33_Appendix}  \end{subfigure}
 & \begin{subfigure}[c]{0.24\textwidth}  \includegraphics[width=\textwidth]{im/RASTA_small01_modif/mixed5b_5x5_pre_reluConv2D_82_RASTA_small01_modif_Deco_toRGB} \caption{} \label{fig:HighFilters_Vizu_RASTA_5b_82_Appendix}  \end{subfigure} \\
     \end{tabular}
     }
      \end{center}
\caption{Optimized Image from different high-level layers. First row InceptionV1 pretrained on ImagneNet, second row fine-tuned on RASTA  \protect\cite{lecoutre_recognizing_2017}.}   
\label{fig:HighLevelFiltersRASTA_Appendix}
 \end{figure}

\begin{figure}[ht!]
     \begin{center}
     \begin{tabular}{ cccc  }
\begin{minipage}[c]{0.24\textwidth} \centering Channel Name :   \end{minipage}  
 & \begin{minipage}[c]{0.24\textwidth} \centering mixed5b\_pool  \_reduce\_pre\_relu:92   \end{minipage}
 & \begin{minipage}[c]{0.24\textwidth} \centering mixed5b\_3x3\_pre\_relu:33  \end{minipage}
 &  \begin{minipage}[c]{0.24\textwidth}\centering mixed5b\_5x5\_pre\_relu:82   \end{minipage} \\
\begin{minipage}[c]{0.24\textwidth} \centering  Imagenet  Pretrained Maximal Activation Examples  \end{minipage}  &
  \begin{subfigure}[c]{0.24\textwidth}      \includegraphics[width=\textwidth]{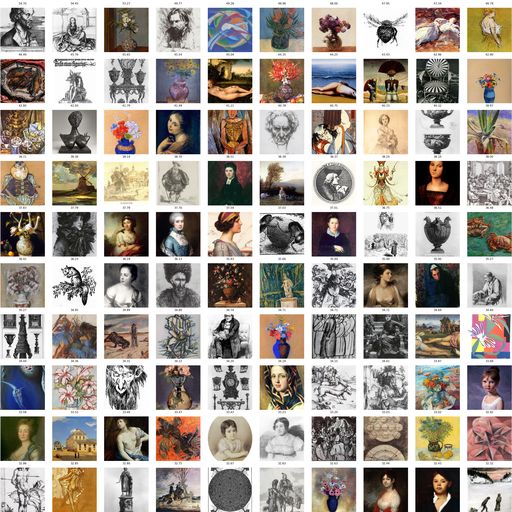}\caption{} \label{fig:HighFilters_Top100_ImageNet_5b_92_Appendix} \end{subfigure}
 &\begin{subfigure}[c]{0.24\textwidth}      \includegraphics[width=\textwidth]{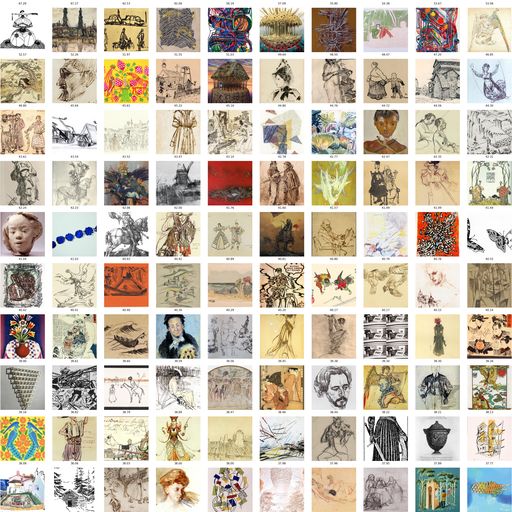} \caption{} \label{fig:HighFilters_Top100_ImageNet_5b_33_Appendix}  \end{subfigure}
 & \begin{subfigure}[c]{0.24\textwidth}   \includegraphics[width=\textwidth]{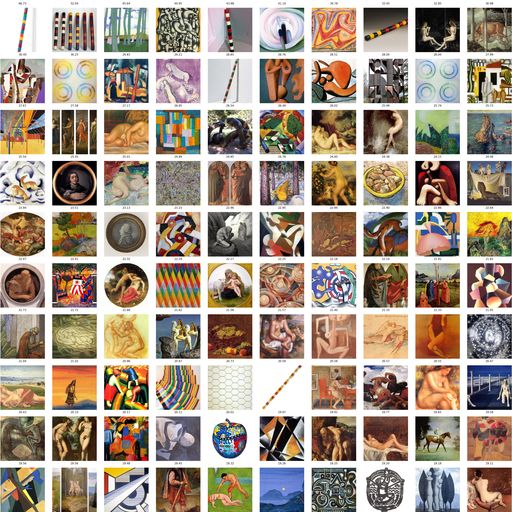} \caption{} \label{fig:HighFilters_Top100_ImageNet_5b_82_Appendix}  \end{subfigure} \\
  \begin{minipage}[c]{0.24\textwidth} \centering Top 100 composition :  \end{minipage}    &    \begin{minipage}[c]{0.24\textwidth} 
     \scriptsize{
 Realism 17\% \\
 Post-Impressionism 10\% \\
 Neoclassicism 10\%  }
 \end{minipage}    &
  \begin{minipage}[c]{0.24\textwidth}
     \scriptsize{
Art\_Nouveau\_(Modern) 14  \% \\
Expressionism 10 \% \\
Realism 9  \% }
 \end{minipage}   & 
      \begin{minipage}[c]{0.24\textwidth}
          \scriptsize{
Surrealism 13\%  \\
Cubism 11\%  \\
Abstract\_Art 10\%  } 
 \end{minipage}     \\
 & & & \\
 \begin{minipage}[c]{0.24\textwidth} \centering   RASTA Fine Tuned Maximal Activation Examples \end{minipage}  &
  \begin{subfigure}[c]{0.24\textwidth}         \includegraphics[width=\textwidth]{im/RASTA_small01_modif/ActivationsImages/RASTA_mixed5b_pool_reduce_pre_relu_92_Most_Pos_Images_NumberIm100_meanAfterRelu} \caption{} \label{fig:HighLevel_Ukiyoe_Appendix} \end{subfigure}
 &\begin{subfigure}[c]{0.24\textwidth}      \includegraphics[width=\textwidth]{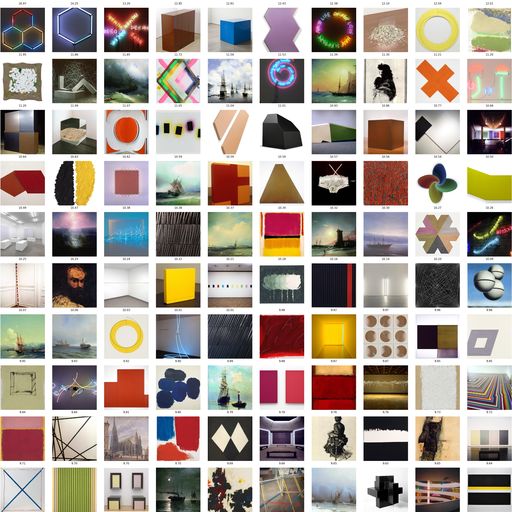}  \caption{} \label{fig:HighLevel_AbstractArt_Appendix}  \end{subfigure}
 & \begin{subfigure}[c]{0.24\textwidth}  
    \includegraphics[width=\textwidth]{im/RASTA_small01_modif/ActivationsImages/RASTA_mixed5b_5x5_pre_relu_82_Most_Pos_Images_NumberIm100_meanAfterRelu} \caption{} \label{fig:HighLevel_Renaissance_Appendix}  \end{subfigure} \\
  \begin{minipage}[c]{0.24\textwidth} \centering Top 100 composition :  \end{minipage}    &    \begin{minipage}[c]{0.24\textwidth}
     \scriptsize{
 Ukiyo-e 82 \%\\
 Northern\_Renaissance 14 \% \\
 Early\_Renaissance 3 \% }
 \end{minipage}  &  \begin{minipage}[c]{0.24\textwidth}
         \scriptsize{
Minimalism 65\% \\
Romanticism 15\% \\
Color\_Field\_Painting 11\%}
\end{minipage}  &  \begin{minipage}[c]{0.24\textwidth}
  \scriptsize{
 Early\_Renaissance 48\%  \\
 High\_Renaissance 27\% \\
 Mannerism\_(Late\_Renaissance) 12\%}
 \end{minipage} \\
Overlapping : & 1\% & 0\% & 0\% \\
     \end{tabular}
      \end{center}
\caption{Maximal Activation Examples for a given channel corresponding to \cref{fig:HighLevelFiltersRASTA_Appendix}. First row InceptionV1 pretrained on ImagneNet, second row fine-tuned on RASTA. The percentage of the 3 most common class is displayed under the images. The percentage of overlapping between the two sets of maximal activation images is displayed at the bottom of each column.}   
\label{fig:HighLevelFiltersRASTA_Top100Im_Appendix}
 \end{figure}

\paragraph{High-level layers.} 
Optimized Images and Maximal activation images for high-level layers can be found in \cref{fig:HighLevelFiltersRASTA_Appendix,fig:HighLevelFiltersRASTA_Top100Im_Appendix}  

\clearpage
\subsection{Training from scratch}
\begin{figure}[ht!]
     \begin{center}
     \begin{tabular}{ ccc  }
\begin{minipage}[c]{0.24\textwidth} \centering Channel Name :   \end{minipage}  
 & \begin{minipage}[c]{0.24\textwidth} \centering mixed4d\_5x5\_pre\_relu:50  \end{minipage}
 & \begin{minipage}[c]{0.24\textwidth} \centering mixed5a\_3x3\_bottleneck\_pre\_relu:1   \end{minipage} \\
 \begin{minipage}[c]{0.24\textwidth} \centering  Initial Partially Random Network Optimized Image \end{minipage}  &
    \begin{subfigure}[c]{0.24\textwidth}         \includegraphics[width=\textwidth]{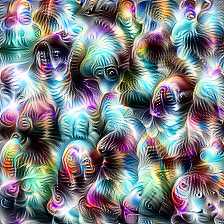}  \caption{}  \label{fig:TheEndFromScratch_50_init_A} \end{subfigure}
 &\begin{subfigure}[c]{0.24\textwidth}         \includegraphics[width=\textwidth]{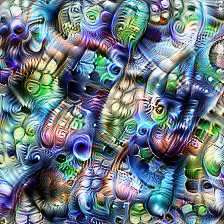}  \caption{}  \label{fig:TheEndFromScratch_1_init_A} \end{subfigure} \\
 
\begin{minipage}[c]{0.24\textwidth} \centering   Initial Partially Random Network 
Maximal Activation Examples \end{minipage}  & 
\begin{subfigure}[c]{0.24\textwidth}   
\includegraphics[width=\textwidth]{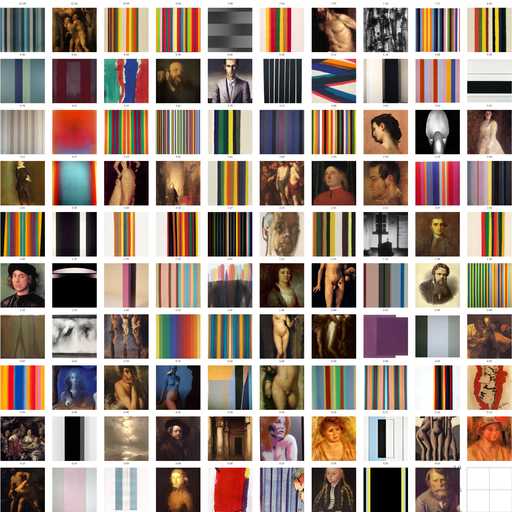}  \caption{ } \label{fig:TheEndFromScratch_Top100_50_init}  \end{subfigure}
 &
  \begin{subfigure}[c]{0.24\textwidth}          \includegraphics[width=\textwidth]{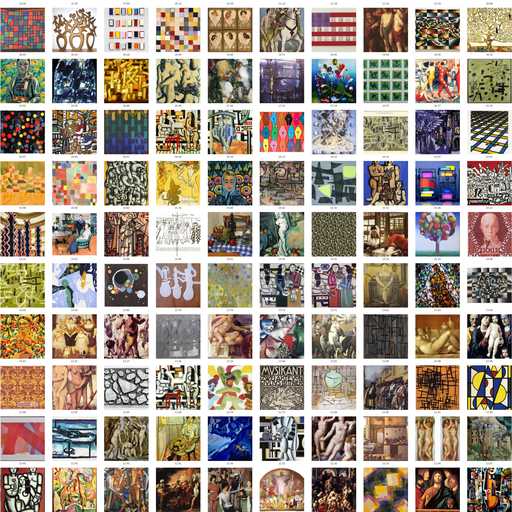}  \caption{ } \label{fig:TheEndFromScratch_Top100_1_init}  \end{subfigure} \\ \\
  \begin{minipage}[c]{0.24\textwidth} \centering Top 100 composition :  \end{minipage}    &    \begin{minipage}[c]{0.24\textwidth}
        \tiny{
Color\_Field\_Painting 41\% \\
Minimalism 12\% \\
Realism 6\% \\}
 \end{minipage}  &  \begin{minipage}[c]{0.24\textwidth}
        \tiny{
Cubism 21.0\% \\
Abstract_Expressionism 10\% \\
Abstract_Art 8\% \\}
\end{minipage}  \\
  \begin{minipage}[c]{0.24\textwidth} \centering  RASTA Fine Tuned Optimized Image \end{minipage}  &
   \begin{subfigure}[c]{0.24\textwidth}         \includegraphics[width=\textwidth]{im/RASTA_big0001_modif_adam_unfreeze50_RandForUnfreezed_SmallDataAug_ep200/mixed4d_5x5_pre_reluConv2D_50_RASTA_big0001_modif_adam_unfreeze50_RandForUnfreezed_SmallDataAug_ep200_Deco_toRGB}  \caption{}  \label{fig:AdaptationFiltersRASTA_checkboard_A} \end{subfigure}
 &\begin{subfigure}[c]{0.24\textwidth}         \includegraphics[width=\textwidth]{im/RASTA_big0001_modif_adam_unfreeze50_RandForUnfreezed_SmallDataAug_ep200/mixed5a_3x3_bottleneck_pre_reluConv2D_1_RASTA_big0001_modif_adam_unfreeze50_RandForUnfreezed_SmallDataAug_ep200_Deco_toRGB}  \caption{}  \label{fig:AdaptationFiltersRASTA_draperie_A} \end{subfigure} \\
 \begin{minipage}[c]{0.24\textwidth} \centering   RASTA Fine Tuned  Maximal Activation Examples  \end{minipage}  &
 \begin{subfigure}[c]{0.24\textwidth}    
\includegraphics[width=\textwidth]{im/RASTA_big0001_modif_adam_unfreeze50_RandForUnfreezed_SmallDataAug_ep200/ActivationsImages/RASTA_mixed4d_5x5_pre_relu_50_Most_Pos_Images_NumberIm100}  \caption{ } \label{fig:RandUnfreeze_Lines_A}  \end{subfigure}
 &
  \begin{subfigure}[c]{0.24\textwidth}          \includegraphics[width=\textwidth]{im/RASTA_big0001_modif_adam_unfreeze50_RandForUnfreezed_SmallDataAug_ep200/ActivationsImages/RASTA_mixed5a_3x3_bottleneck_pre_relu_1_Most_Pos_Images_NumberIm100}  \caption{ } \label{fig:RandUnfreeze_Renaissance_A}  \end{subfigure} \\
  \begin{minipage}[c]{0.24\textwidth} \centering Top 100 composition :  \end{minipage}    &    \begin{minipage}[c]{0.24\textwidth}
        \tiny{
Abstract\_Expressionism 24\% \\
Minimalism 13\% \\
Art\_Informel 9\%}
 \end{minipage}  &  \begin{minipage}[c]{0.24\textwidth}
        \tiny{
 Northern\_Renaissance 39\% \\
Romanticism 20\% \\
Early_Renaissance 18\% \\}
\end{minipage}  \\
Overlapping : & 0\% & 0\%  \\
     \end{tabular}
      \end{center}
\vspace*{-5mm}
\caption{Optimized Image and Maximal activation examples for the model fine-tuned with low-level layers initialized from ImagNet and upper layers initialized at random. First row: optimized images for the initial partially random model. Second row: top 100 maximal activation examples for the same model. Third and fourth rows: optimized images and maximal activation examples for the same channel of the model trained from scratch. The percentage of the 3 most common class is displayed under the images. The overlapping ratio between Top 100 maximal activation images is printed on the last line.}   
\label{fig:TheEndFromScratch_appendix}
 \end{figure}

\begin{figure}[ht!]
     \begin{center}
     \resizebox{\columnwidth}{!}{
     \begin{tabular}{ cccc  }
\begin{minipage}[c]{0.24\textwidth} \centering Channel Name :   \end{minipage}  
 & \begin{minipage}[c]{0.24\textwidth} \centering mixed4d:8  \end{minipage}
 & \begin{minipage}[c]{0.24\textwidth} \centering mixed4d:16 \end{minipage}
 &  \begin{minipage}[c]{0.24\textwidth}\centering mixed4d:66  \end{minipage} \\
 \begin{minipage}[c]{0.24\textwidth} \centering  Initial Random Network Optimized Image \end{minipage}  &
  \begin{subfigure}[c]{0.24\textwidth}      \includegraphics[width=\textwidth]{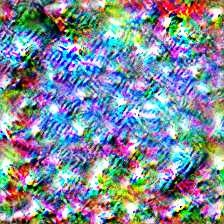} \caption{} \label{fig:LearnedFromScratch_8_init_A} \end{subfigure}
 &\begin{subfigure}[c]{0.24\textwidth}   \includegraphics[width=\textwidth]{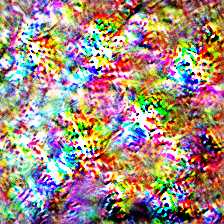}  \caption{} \label{fig:LearnedFromScratch_16_init_A}  \end{subfigure}
 & \begin{subfigure}[c]{0.24\textwidth}  \includegraphics[width=\textwidth]{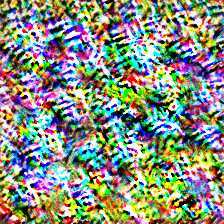} \caption{} \label{fig:LearnedFromScratch_66_init_A}  \end{subfigure} \\
 
\begin{minipage}[c]{0.24\textwidth} \centering  Initial Random Network 
Maximal Activation Examples \end{minipage}  &
  \begin{subfigure}[c]{0.24\textwidth}     \includegraphics[width=\textwidth]{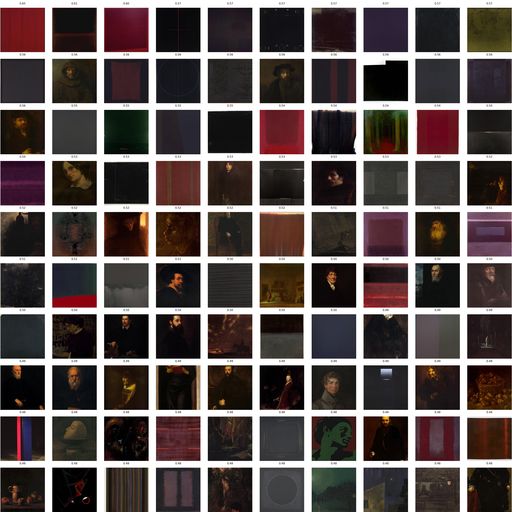}  \caption{} \label{fig:LearnedFromScratch_Top100_8_Init}  \end{subfigure}
 &\begin{subfigure}[c]{0.24\textwidth}      \includegraphics[width=\textwidth]{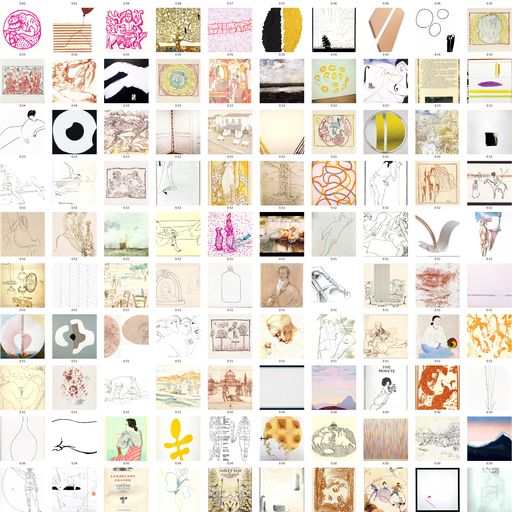} \caption{} \label{fig:LearnedFromScratch_Top100_16_Init}  \end{subfigure}
 & \begin{subfigure}[c]{0.24\textwidth}    \includegraphics[width=\textwidth]{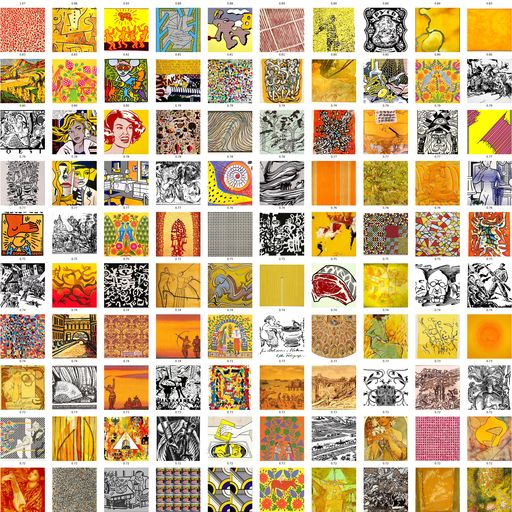} \caption{} \label{fig:LearnedFromScratch_Top100_66_Init}  \end{subfigure} \\
 \\
  \begin{minipage}[c]{0.24\textwidth} \centering Top 100 composition :  \end{minipage}    &    \begin{minipage}[c]{0.24\textwidth}
        \tiny{
Minimalism 19\% \\
Realism 11\% \\
Baroque 10\% \\}
 \end{minipage}  &  \begin{minipage}[c]{0.24\textwidth}
        \tiny{
Expressionism 15\% \\
Minimalism 12\% \\
Art\_Nouveau\_(Modern) 11.0\% \\}
\end{minipage}  &  \begin{minipage}[c]{0.24\textwidth}
        \tiny{
Pop\_Art 17\% \\
Na\"ive\_Art\_(Primitivism) 15\% \\
Abstract\_Art 9\% \\}
 \end{minipage} \\
  \begin{minipage}[c]{0.24\textwidth} \centering  RASTA Fine Tuned Optimized Image \end{minipage}  &
  \begin{subfigure}[c]{0.24\textwidth}      \includegraphics[width=\textwidth]{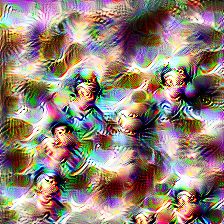} \caption{} \label{fig:LearnedFromScratch_8_A} \end{subfigure}
 &\begin{subfigure}[c]{0.24\textwidth}   \includegraphics[width=\textwidth]{im/RASTA_big001_modif_RandInit_randomCrop_deepSupervision_ep200_LRschedG/mixed4dRelu_16_RASTA_big001_modif_RandInit_randomCrop_deepSupervision_ep200_LRschedG_Deco_toRGB}  \caption{} \label{fig:LearnedFromScratch_Ukiyoe_A}  \end{subfigure}
 & \begin{subfigure}[c]{0.24\textwidth}  \includegraphics[width=\textwidth]{im/RASTA_big001_modif_RandInit_randomCrop_deepSupervision_ep200_LRschedG/mixed4dRelu_66_RASTA_big001_modif_RandInit_randomCrop_deepSupervision_ep200_LRschedG_Deco_toRGB} \caption{} \label{fig:LearnedFromScratch_Magic_A}  \end{subfigure} \\
 \begin{minipage}[c]{0.24\textwidth} \centering   RASTA Fine Tuned  Maximal Activation Examples  \end{minipage}  &
  \begin{subfigure}[c]{0.24\textwidth}      \includegraphics[width=\textwidth]{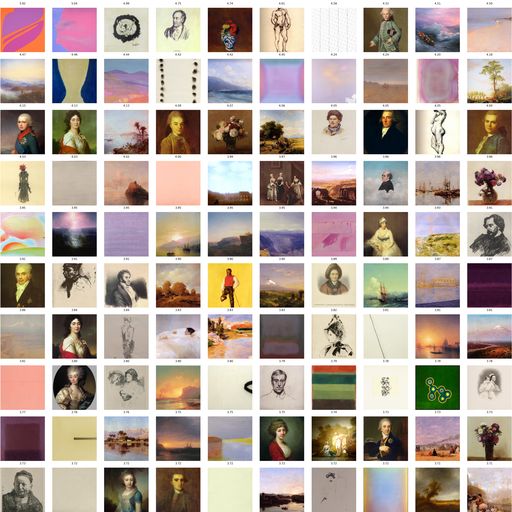}\caption{} \label{fig:LearnedFromScratch_Top100_8_A} \end{subfigure}
 &\begin{subfigure}[c]{0.24\textwidth}          \includegraphics[width=\textwidth]{im/RASTA_big001_modif_RandInit_randomCrop_deepSupervision_ep200_LRschedG/ActivationsImages/RASTA_mixed4d_16_Most_Pos_Images_NumberIm100}  \caption{} \label{fig:LearnedFromScratch_Top100_Ukiyoe_A}  \end{subfigure}
 & \begin{subfigure}[c]{0.24\textwidth}  
    \includegraphics[width=\textwidth]{im/RASTA_big001_modif_RandInit_randomCrop_deepSupervision_ep200_LRschedG/ActivationsImages/RASTA_mixed4d_66_Most_Pos_Images_NumberIm100}\caption{} \label{fig:LearnedFromScratch_Top100_Magic_A}  \end{subfigure} \\
  \begin{minipage}[c]{0.24\textwidth} \centering Top 100 composition :  \end{minipage}    &    \begin{minipage}[c]{0.24\textwidth}
        \tiny{
Romanticism 23\% \\
Rococo 13\% \\
Realism 13\% \\}
 \end{minipage}  &  \begin{minipage}[c]{0.24\textwidth}
        \tiny{
Ukiyo-e 85\% \\
Art\_Nouveau\_(Modern) 11\% \\
Northern\_Renaissance 2\% \\}
\end{minipage}  &  \begin{minipage}[c]{0.24\textwidth}
        \tiny{
Magic\_Realism 78\% \\
Ukiyo-e 22\% \\}
 \end{minipage} \\
 Overlapping : & 0\% & 0\% & 0\% \\
 \end{tabular}
 }
 \end{center}
\caption{First row: optimized images for the initial random model.  Second row: top 100 maximal activation examples for the same model. Third and fourth rows: optimized images and maximal activation examples for the same channel of the model trained from scratch. The percentage of the 3 most common class is displayed under the images. The overlapping ratio between Top 100 maximal activation images is printed on the last line.}   
\label{fig:LearnedFromScratch_Top100_appendix}
 \end{figure}

More complete example of optimized images and maximal activation examples are displayed in \cref{fig:TheEndFromScratch_appendix,fig:LearnedFromScratch_Top100_appendix}.
 
\clearpage
\subsection{Quantitative evaluations.}
\begin{figure}[!tbp]
  \centering
  \begin{subfigure}[t]{0.49\textwidth}
  \resizebox{\textwidth}{!}{
\begin{tikzpicture}

\definecolor{color0}{rgb}{0.894117647058824,0.101960784313725,0.109803921568627}
\definecolor{color1}{rgb}{0.968627450980392,0.505882352941176,0.749019607843137}
\definecolor{color2}{rgb}{0.650980392156863,0.337254901960784,0.156862745098039}
\definecolor{color3}{rgb}{0.870588235294118,0.870588235294118,0}
\definecolor{color4}{rgb}{1,0.737254901960784,0.474509803921569}
\definecolor{color5}{rgb}{0.454901960784314,0.498039215686275,0.729411764705882}
\definecolor{color6}{rgb}{0.866666666666667,0.517647058823529,0.32156862745098}

\begin{axis}[
axis line style={white!80!black},
tick align=outside,
tick pos=both,
x grid style={white!80!black},
xlabel={Layer},
xmajorgrids,
xmin=0.5, xmax=7.5,
xtick style={color=white!15!black},
xtick={1,2,3,4,5,6,7},
xticklabel style = {rotate=45.0},
xticklabels={mixed4a,mixed4b,mixed4c,mixed4d,mixed4e,mixed5a,mixed5b},
y grid style={white!80!black},
ylabel={Overlapping (\%)},
ymajorgrids,
ymin=-3.9, ymax=81.9,
ytick style={color=white!15!black}
]
\path [draw=white, fill=color0, opacity=0.5]
(axis cs:0.75,14.75)
--(axis cs:1.25,14.75)
--(axis cs:1.25,50)
--(axis cs:0.75,50)
--cycle;
\path [draw=white, fill=color1, opacity=0.5]
(axis cs:1.75,4)
--(axis cs:2.25,4)
--(axis cs:2.25,23)
--(axis cs:1.75,23)
--cycle;
\path [draw=white, fill=color2, opacity=0.5]
(axis cs:2.75,1)
--(axis cs:3.25,1)
--(axis cs:3.25,7.25)
--(axis cs:2.75,7.25)
--cycle;
\path [draw=white, fill=color3, opacity=0.5]
(axis cs:3.75,0)
--(axis cs:4.25,0)
--(axis cs:4.25,3)
--(axis cs:3.75,3)
--cycle;
\path [draw=white, fill=color4, opacity=0.5]
(axis cs:4.75,0)
--(axis cs:5.25,0)
--(axis cs:5.25,2)
--(axis cs:4.75,2)
--cycle;
\path [draw=white, fill=white!60!black, opacity=0.5]
(axis cs:5.75,0)
--(axis cs:6.25,0)
--(axis cs:6.25,1)
--(axis cs:5.75,1)
--cycle;
\path [draw=white, fill=color5, opacity=0.5]
(axis cs:6.75,0)
--(axis cs:7.25,0)
--(axis cs:7.25,0)
--(axis cs:6.75,0)
--cycle;
\addplot [black]
table {%
0.75 14.75
1.25 14.75
1.25 50
0.75 50
0.75 14.75
};
\addplot [black]
table {%
1 14.75
1 0
};
\addplot [black]
table {%
1 50
1 78
};
\addplot [black]
table {%
0.875 0
1.125 0
};
\addplot [black]
table {%
0.875 78
1.125 78
};
\addplot [black]
table {%
1.75 4
2.25 4
2.25 23
1.75 23
1.75 4
};
\addplot [black]
table {%
2 4
2 0
};
\addplot [black]
table {%
2 23
2 51
};
\addplot [black]
table {%
1.875 0
2.125 0
};
\addplot [black]
table {%
1.875 51
2.125 51
};
\addplot [black, mark=+, mark size=3, mark options={solid,fill opacity=0}, only marks]
table {%
2 59
2 58
2 59
2 60
};
\addplot [black]
table {%
2.75 1
3.25 1
3.25 7.25
2.75 7.25
2.75 1
};
\addplot [black]
table {%
3 1
3 0
};
\addplot [black]
table {%
3 7.25
3 16
};
\addplot [black]
table {%
2.875 0
3.125 0
};
\addplot [black]
table {%
2.875 16
3.125 16
};
\addplot [black, mark=+, mark size=3, mark options={solid,fill opacity=0}, only marks]
table {%
3 26
3 21
3 19
3 19
3 35
3 17
3 20
3 17
3 22
3 20
3 28
3 18
3 33
3 29
3 27
3 31
3 37
3 19
3 28
3 27
3 20
3 37
3 17
3 17
3 23
3 19
3 28
3 17
3 33
3 21
3 17
3 20
3 17
3 21
3 25
3 39
3 25
3 23
3 21
3 19
3 19
3 21
3 20
};
\addplot [black]
table {%
3.75 0
4.25 0
4.25 3
3.75 3
3.75 0
};
\addplot [black]
table {%
4 0
4 0
};
\addplot [black]
table {%
4 3
4 7
};
\addplot [black]
table {%
3.875 0
4.125 0
};
\addplot [black]
table {%
3.875 7
4.125 7
};
\addplot [black, mark=+, mark size=3, mark options={solid,fill opacity=0}, only marks]
table {%
4 12
4 27
4 10
4 10
4 10
4 8
4 13
4 35
4 20
4 9
4 8
4 8
4 16
4 16
4 16
4 8
4 8
4 9
4 11
4 18
4 8
4 17
4 8
4 8
4 16
4 29
4 9
4 9
4 8
4 9
4 8
4 9
4 9
4 10
4 12
4 9
4 11
4 15
4 17
4 9
4 11
4 8
4 10
};
\addplot [black]
table {%
4.75 0
5.25 0
5.25 2
4.75 2
4.75 0
};
\addplot [black]
table {%
5 0
5 0
};
\addplot [black]
table {%
5 2
5 5
};
\addplot [black]
table {%
4.875 0
5.125 0
};
\addplot [black]
table {%
4.875 5
5.125 5
};
\addplot [black, mark=+, mark size=3, mark options={solid,fill opacity=0}, only marks]
table {%
5 6
5 6
5 8
5 12
5 7
5 10
5 6
5 9
5 8
5 8
5 6
5 7
5 8
5 6
5 9
5 13
5 12
5 6
5 7
5 6
5 8
5 6
5 8
5 12
5 39
5 9
5 14
5 9
5 8
5 8
5 6
5 7
5 8
5 8
5 11
5 8
5 7
5 9
5 13
5 7
5 10
5 6
5 6
5 8
5 8
5 11
5 10
5 12
5 6
5 6
5 6
5 8
};
\addplot [black]
table {%
5.75 0
6.25 0
6.25 1
5.75 1
5.75 0
};
\addplot [black]
table {%
6 0
6 0
};
\addplot [black]
table {%
6 1
6 2
};
\addplot [black]
table {%
5.875 0
6.125 0
};
\addplot [black]
table {%
5.875 2
6.125 2
};
\addplot [black, mark=+, mark size=3, mark options={solid,fill opacity=0}, only marks]
table {%
6 3
6 5
6 6
6 5
6 8
6 9
6 3
6 4
6 5
6 4
6 3
6 3
6 5
6 3
6 4
6 3
6 4
6 4
6 3
6 11
6 3
6 6
6 3
6 6
6 3
6 3
6 4
6 4
6 4
6 3
6 4
6 5
6 3
6 11
6 4
6 3
6 3
6 3
6 6
6 3
6 5
6 3
6 17
6 3
6 3
6 3
6 7
6 3
6 3
6 4
6 3
6 7
6 6
6 3
6 4
6 5
6 3
6 6
6 3
6 6
6 6
6 3
6 3
6 3
6 4
6 3
6 5
6 5
6 5
6 14
6 4
6 7
6 8
6 4
6 13
6 4
6 11
6 9
6 8
6 11
};
\addplot [black]
table {%
6.75 0
7.25 0
7.25 0
6.75 0
6.75 0
};
\addplot [black]
table {%
7 0
7 0
};
\addplot [black]
table {%
7 0
7 0
};
\addplot [black]
table {%
6.875 0
7.125 0
};
\addplot [black]
table {%
6.875 0
7.125 0
};
\addplot [black, mark=+, mark size=3, mark options={solid,fill opacity=0}, only marks]
table {%
7 1
7 3
7 1
7 3
7 6
7 1
7 2
7 2
7 4
7 4
7 1
7 4
7 11
7 5
7 1
7 1
7 1
7 5
7 2
7 1
7 4
7 3
7 1
7 1
7 1
7 5
7 2
7 2
7 1
7 2
7 1
7 3
7 3
7 6
7 1
7 1
7 1
7 5
7 1
7 1
7 1
7 2
7 1
7 4
7 1
7 2
7 2
7 1
7 12
7 2
7 2
7 1
7 1
7 1
7 1
7 2
7 17
7 1
7 4
7 1
7 3
7 1
7 1
7 2
7 2
7 2
7 2
7 1
7 1
7 2
7 1
7 1
7 1
7 1
7 1
7 1
7 1
7 1
7 1
7 3
7 1
7 6
7 1
7 1
7 10
7 1
7 1
7 2
7 1
7 1
7 5
7 2
7 6
7 2
7 1
7 1
7 2
7 1
7 3
7 5
7 1
7 2
7 2
7 1
7 1
7 1
7 2
7 1
7 2
7 1
7 1
7 1
7 2
7 2
7 1
7 3
7 3
7 1
7 1
7 5
7 1
7 2
7 8
7 1
7 1
7 5
7 2
7 2
7 1
7 2
7 1
7 2
7 1
7 1
7 1
7 1
7 1
7 1
7 1
7 1
7 2
7 2
7 1
7 3
7 1
7 1
7 1
7 1
7 1
7 2
7 2
7 5
7 1
7 1
7 1
7 1
7 1
7 1
7 4
7 1
7 1
7 2
7 1
7 4
7 1
7 2
7 2
7 1
7 1
7 1
7 1
7 1
7 3
7 2
7 2
7 1
7 1
7 2
7 1
7 1
7 2
7 1
7 3
7 1
7 1
7 1
7 1
7 1
7 1
7 1
7 1
7 1
7 2
7 3
7 2
7 3
7 1
7 4
7 2
7 1
7 2
7 3
7 1
7 1
7 1
7 8
7 1
7 1
7 1
7 1
7 2
7 1
7 2
7 2
7 1
7 1
7 1
7 1
7 1
7 1
7 2
7 1
7 1
7 3
7 4
};
\addplot [semithick, white!10!black]
table {%
0.75 29
};
\addplot [semithick, white!10!black]
table {%
0.75 29
1.25 29
};
\addplot [semithick, white, mark=star, mark size=3, mark options={solid,draw=white!10!black}]
table {%
1 32.1751968503937
};
\addplot [semithick, white!10!black]
table {%
1.75 11
};
\addplot [semithick, white!10!black]
table {%
1.75 11
2.25 11
};
\addplot [semithick, white, mark=star, mark size=3, mark options={solid,draw=white!10!black}]
table {%
2 14.189453125
};
\addplot [semithick, white!10!black]
table {%
2.75 3
};
\addplot [semithick, white!10!black]
table {%
2.75 3
3.25 3
};
\addplot [semithick, white, mark=star, mark size=3, mark options={solid,draw=white!10!black}]
table {%
3 5.619140625
};
\addplot [semithick, white!10!black]
table {%
3.75 1
};
\addplot [semithick, white!10!black]
table {%
3.75 1
4.25 1
};
\addplot [semithick, white, mark=star, mark size=3, mark options={solid,draw=white!10!black}]
table {%
4 2.34469696969697
};
\addplot [semithick, white!10!black]
table {%
4.75 0
};
\addplot [semithick, white!10!black]
table {%
4.75 0
5.25 0
};
\addplot [semithick, white, mark=star, mark size=3, mark options={solid,draw=white!10!black}]
table {%
5 1.33774038461538
};
\addplot [semithick, white!10!black]
table {%
5.75 0
};
\addplot [semithick, white!10!black]
table {%
5.75 0
6.25 0
};
\addplot [semithick, white, mark=star, mark size=3, mark options={solid,draw=white!10!black}]
table {%
6 0.801682692307692
};
\addplot [semithick, white!10!black]
table {%
6.75 0
};
\addplot [semithick, white!10!black]
table {%
6.75 0
7.25 0
};
\addplot [semithick, white, mark=star, mark size=3, mark options={solid,draw=white!10!black}]
table {%
7 0.4453125
};
\addplot [color6]
table {%
0.75 29
1.25 29
};
\addplot [color6]
table {%
1.75 11
2.25 11
};
\addplot [color6]
table {%
2.75 3
3.25 3
};
\addplot [color6]
table {%
3.75 1
4.25 1
};
\addplot [color6]
table {%
4.75 0
5.25 0
};
\addplot [color6]
table {%
5.75 0
6.25 0
};
\addplot [color6]
table {%
6.75 0
7.25 0
};
\end{axis}

\end{tikzpicture}
}
    \caption{Model fine-tuned on RASTA with low-level layers initialized from ImagNet and upper layers initialized at random.} \label{fig:100_Boxplots_per_layer_overlap_RandForUnfreeze}
  \end{subfigure}  
  \begin{subfigure}[t]{0.49\textwidth}
  \resizebox{\textwidth}{!}{
\input{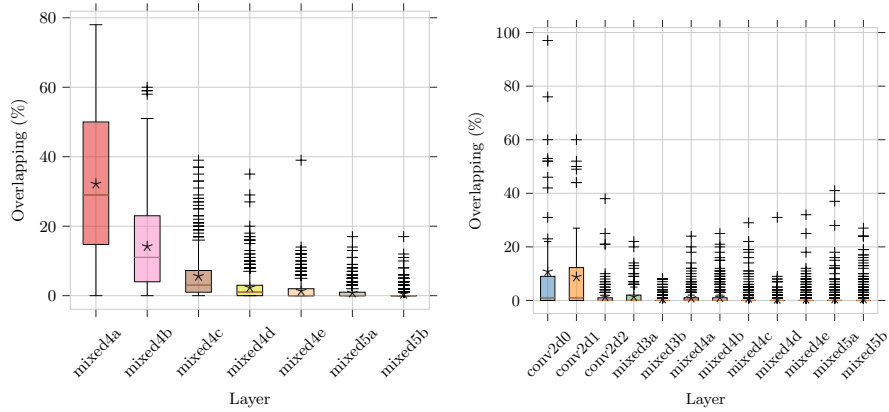}
}
    \caption{Model trained from scratch on RASTA.} \label{fig:100_Boxplots_per_layer_overlap_fromScratch}
  \end{subfigure}  
 \caption{Boxplots of the overlapping ratio on the top 100 maximal activation images for the two different models trained from scratch (between the random initialization and the trained model).
  For each box, the horizontal orange line corresponds to the average result and the star to the median.}
  \label{fig:100_Boxplots_per_layer_overlap_A}
\end{figure}

\begin{figure}[!tbp]
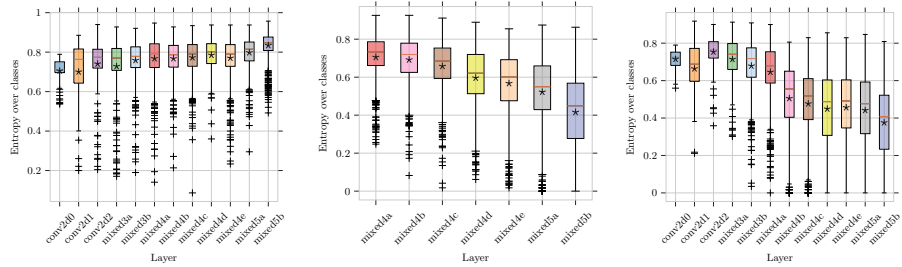

  \centering
  \begin{subfigure}[t]{0.32\textwidth}
  \resizebox{\textwidth}{!}{
\input{imInTex/pretrained/Overlapping/Purity_entropy100_Boxplots_per_layer.tex}
}
    \caption{ImageNet pretained.} \label{fig:entropy_Boxplots_per_layer_overlap_pretrained}
  \end{subfigure}    
      \hfill
  \begin{subfigure}[t]{0.32\textwidth}
  \resizebox{\textwidth}{!}{
\input{imInTex/RASTA_big0001_modif_adam_unfreeze50_RandForUnfreezed_SmallDataAug_ep200/Overlapping/Purity_entropy100_Boxplots_per_layer.tex}
}
    \caption{Model fine-tuned on RASTA with low-level layers initialized from ImagNet and upper layers initialized at random.} \label{fig:entropy_Boxplots_per_layer_overlap_RandForUnfreeze}
  \end{subfigure}  
  \hfill
  \begin{subfigure}[t]{0.32\textwidth}
  \resizebox{\textwidth}{!}{
\input{imInTex/RASTA_big001_modif_RandInit_randomCrop_deepSupervision_ep200_LRschedG/Overlapping/Purity_entropy100_Boxplots_per_layer.tex}
}
    \caption{Model trained from scratch on RASTA.} \label{fig:entropy_Boxplots_per_layer_overlap_fromScratch}
  \end{subfigure}
    \hfill
 \caption{Boxplots of the entropy over classes on the top 100 maximal activation images for the two different models trained from scratch.
  For each box, the horizontal orange line corresponds to the average result and the star to the median.}
  \label{fig:100_Boxplots_per_layer_entropy_A}
\end{figure}

Boxplots of some metrics on the top 100 maximal activation images for other models can be found in \cref{fig:100_Boxplots_per_layer_overlap_A,fig:100_Boxplots_per_layer_entropy_A}.

\subsection{From One Art dataset to another}
\begin{figure}[ht!]
     \begin{center}
    \resizebox{\columnwidth}{!}{
     \begin{tabular}{ cccc  }
\begin{minipage}[c]{0.24\textwidth} \centering Channel Name :   \end{minipage}  
 & \begin{minipage}[c]{0.24\textwidth} \centering mixed4c\_pool \\ \_reduce\_pre\_relu:2 \end{minipage}
 & \begin{minipage}[c]{0.24\textwidth} \centering mixed4c\_3x3 \\ \_bottleneck\_pre\_relu:78 \end{minipage}
 &  \begin{minipage}[c]{0.24\textwidth}\centering mixed4d\_5x5\_pre\_relu:49  \end{minipage} \\
\begin{minipage}[c]{0.24\textwidth} \centering  Imagenet  Pretrained Optimized Image \end{minipage}  &
  \begin{subfigure}[c]{0.24\textwidth}     \includegraphics[width=\textwidth]{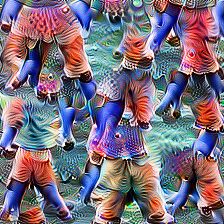}  \caption{} \label{fig:MidLevelFilters_ImageNet_4c_2}  \end{subfigure}
 &\begin{subfigure}[c]{0.24\textwidth}      \includegraphics[width=\textwidth]{im/pretrained/mixed4c_3x3_bottleneck_pre_reluConv2D_78_Imagnet_Deco_toRGB}  \caption{} \label{fig:MidLevelFilters_ImageNet_4c_78}  \end{subfigure}
 & \begin{subfigure}[c]{0.24\textwidth}   \includegraphics[width=\textwidth]{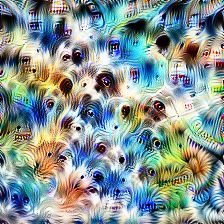} \caption{} \label{fig:MidLevelFilters_ImageNet_4d_49}  \end{subfigure} \\
  \begin{minipage}[c]{0.24\textwidth} \centering   IconArt Fine Tuned  Optimized Image  \end{minipage}  &  \begin{subfigure}[c]{0.24\textwidth}     \includegraphics[width=\textwidth]{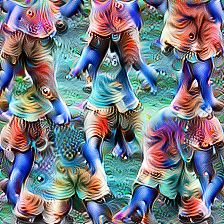}  \caption{} \label{fig:MidLevelFilters_IconArt_4c_2}  \end{subfigure}
 &\begin{subfigure}[c]{0.24\textwidth}      \includegraphics[width=\textwidth]{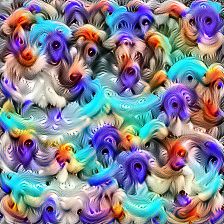}  \caption{} \label{fig:MidLevelFilters_IconArt_4c_78}  \end{subfigure}
 & \begin{subfigure}[c]{0.24\textwidth}   \includegraphics[width=\textwidth]{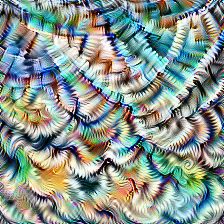} \caption{} \label{fig:MidLevelFilters_IconArt_4d_49}  \end{subfigure} \\
 \begin{minipage}[c]{0.24\textwidth} \centering   RASTA Fine Tuned  Optimized Image  \end{minipage}  &  \begin{subfigure}[c]{0.24\textwidth}     \includegraphics[width=\textwidth]{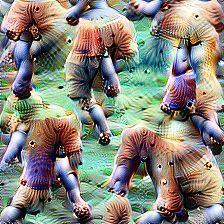}  \caption{} \label{fig:MidLevelFilters_RASTA_4c_2}  \end{subfigure}
 &\begin{subfigure}[c]{0.24\textwidth}      \includegraphics[width=\textwidth]{im/RASTA_small01_modif/mixed4c_3x3_bottleneck_pre_reluConv2D_78_RASTA_small01_modif_Deco_toRGB}  \caption{} \label{fig:MidLevelFilters_RASTA_4c_78}  \end{subfigure}
 & \begin{subfigure}[c]{0.24\textwidth}   \includegraphics[width=\textwidth]{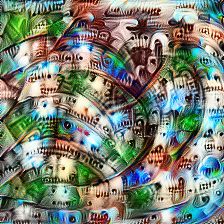} \caption{} \label{fig:MidLevelFilters_RASTA_4d_49}  \end{subfigure} \\
 \begin{minipage}[c]{0.24\textwidth} \centering  RASTA and IconArt Fine Tuned Optimized Image  \end{minipage}  &  \begin{subfigure}[c]{0.24\textwidth}     \includegraphics[width=\textwidth]{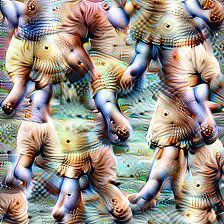}  \caption{} \label{fig:AdaptationFiltersRASTAIconArt_nudity}  \end{subfigure}
 &\begin{subfigure}[c]{0.24\textwidth}      \includegraphics[width=\textwidth]{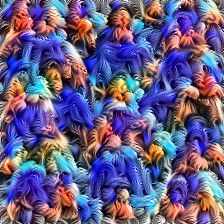}  \caption{} \label{fig:AdaptationFiltersRASTAIconArt_blueDrapery}  \end{subfigure}
 & \begin{subfigure}[c]{0.24\textwidth}   \includegraphics[width=\textwidth]{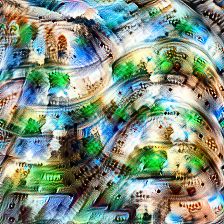} \caption{} \label{fig:AdaptationFiltersRASTAIconArt_treeblue}  \end{subfigure} \\ 
     \end{tabular}
     }
      \end{center}
\caption{Optimized Image for one Individual channel from different mid-level layers. First row InceptionV1 pretrained on ImagneNet, second row fine-tuned on IconArt, third row on RASTA, last row fine-tuned first on RASTA  \protect\cite{lecoutre_recognizing_2017} then on IconArt  \protect\cite{gonthier_weakly_2018}.}   
\label{fig:MidLevelFilters_RASTA_IconArt}
 \end{figure}

The optimized images of the model pretrained on ImageNet, fine-tuned first on RASTA \cite{lecoutre_recognizing_2017} and then on IconArt \cite{gonthier_weakly_2018}, show that some of the filters learned on RASTA can be directly reused  for the IconArt dataset such as a nudity detector (\cref{fig:AdaptationFiltersRASTAIconArt_nudity}). Although this class doesn't exist in the RASTA dataset, images with such visual pattern belong to this dataset. 
Some filters are adapted to focus more on a given visual pattern, for instance the blue drapery detector (\cref{fig:AdaptationFiltersRASTAIconArt_blueDrapery}). This new detector is not learned by the model with only a fine-tuning on IconArt (see \cref{fig:MidLevelFilters_IconArt_4c_78}). 
Other filters are completely changed such as the tree in front of a blue sky detector (\cref{fig:AdaptationFiltersRASTAIconArt_treeblue}). 
The datasets used are not representative of all the artistic production by far and are certainly biased. However these illustrations allow us to highlight what kinds of filters can be relevant for artwork classification.
\begin{figure}[ht!]
    \begin{center}
    \resizebox{\columnwidth}{!}{
    \begin{tabular}{ cccc  }
\begin{minipage}[c]{0.24\textwidth} \centering Channel Name :   \end{minipage}  
& \begin{minipage}[c]{0.24\textwidth} \centering mixed4c\_pool \\ \_reduce\_pre\_relu:2 \end{minipage}
& \begin{minipage}[c]{0.24\textwidth} \centering mixed4c\_3x3 \\ \_bottleneck\_pre\_relu:78 \end{minipage}
&  \begin{minipage}[c]{0.24\textwidth}\centering mixed4d\_5x5\_pre\_relu:49  \end{minipage} \\
\begin{minipage}[c]{0.24\textwidth} \centering  Imagenet  Pretrained Maximal Activation Examples  \end{minipage}  &
 \begin{subfigure}[c]{0.24\textwidth}   \includegraphics[width=\textwidth]{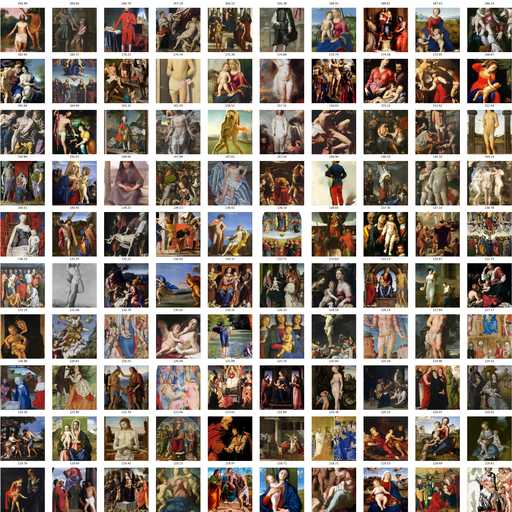}  \caption{} \label{fig:MidLevelFilters_ImageNet_4c_2_Top100}  \end{subfigure}
&\begin{subfigure}[c]{0.24\textwidth}       \includegraphics[width=\textwidth]{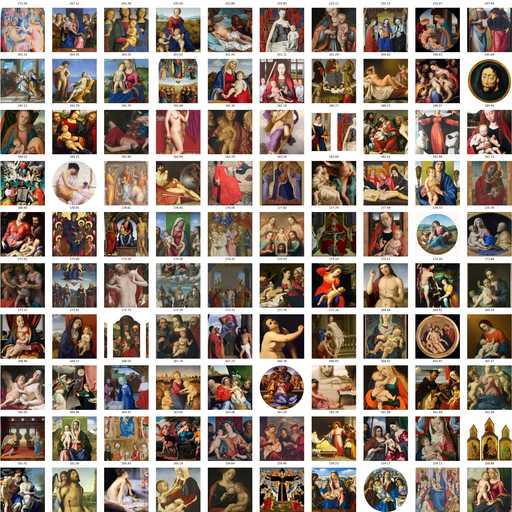}  \caption{} \label{fig:MidLevelFilters_ImageNet_4c_78_Top100}  \end{subfigure}
& \begin{subfigure}[c]{0.24\textwidth}    \includegraphics[width=\textwidth]{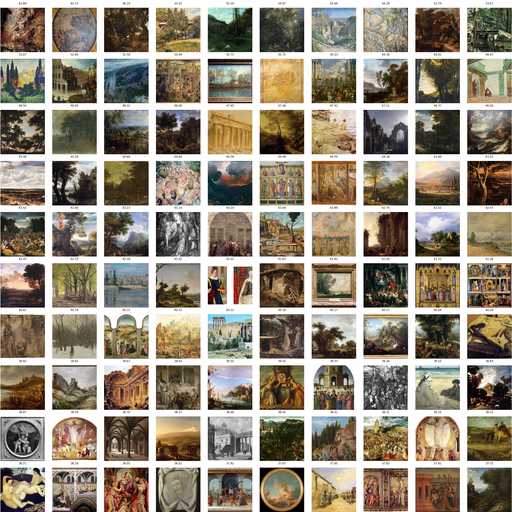}  \caption{} \label{fig:MidLevelFilters_ImageNet_4d_49_Top100}  \end{subfigure} \\
\begin{minipage}[c]{0.24\textwidth} \centering   IconArt Fine Tuned  Maximal Activation Examples   \end{minipage}  &
\begin{subfigure}[c]{0.24\textwidth}      \includegraphics[width=\textwidth]{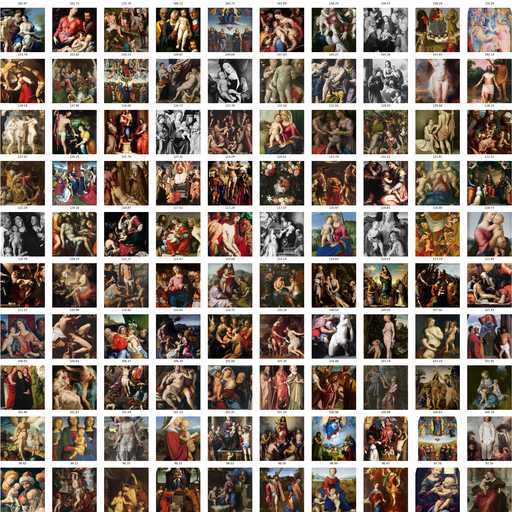}  \caption{} \label{fig:MidLevelFilters_IconArt_4c_2_Top100 }  \end{subfigure} &
\begin{subfigure}[c]{0.24\textwidth}    
  \includegraphics[width=\textwidth]{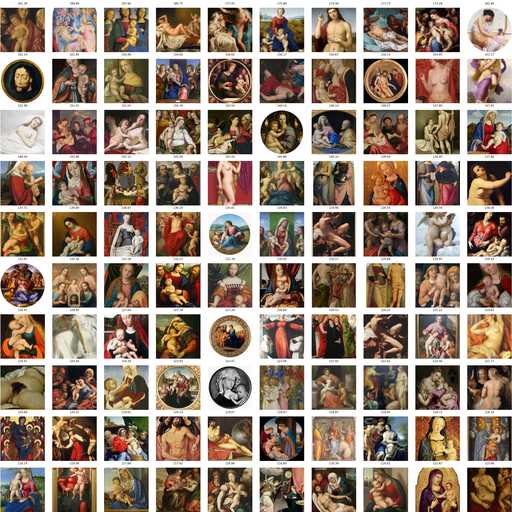}  \caption{} \label{fig:MidLevelFilters_IconArt_4c_78_Top100}  \end{subfigure}
& \begin{minipage}[c]{0.24\textwidth}   
No positive images on this channel.
\end{minipage}
\\
\begin{minipage}[c]{0.24\textwidth} \centering   RASTA Fine Tuned  Maximal Activation Examples   \end{minipage}  &  \begin{subfigure}[c]{0.24\textwidth}    
  \includegraphics[width=\textwidth]{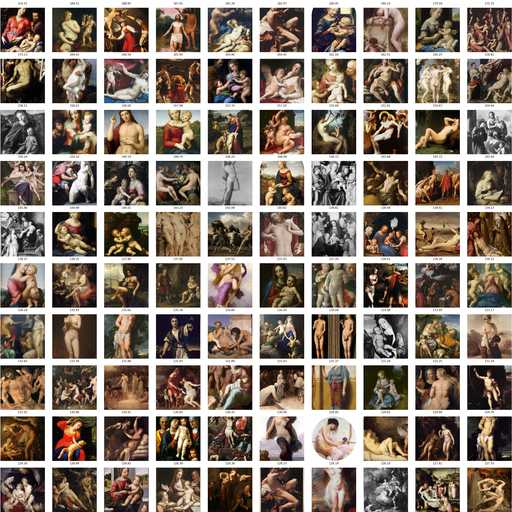}  \caption{} \label{fig:MidLevelFilters_RASTA_4c_2_Top100}  \end{subfigure}
&\begin{subfigure}[c]{0.24\textwidth}      \includegraphics[width=\textwidth]{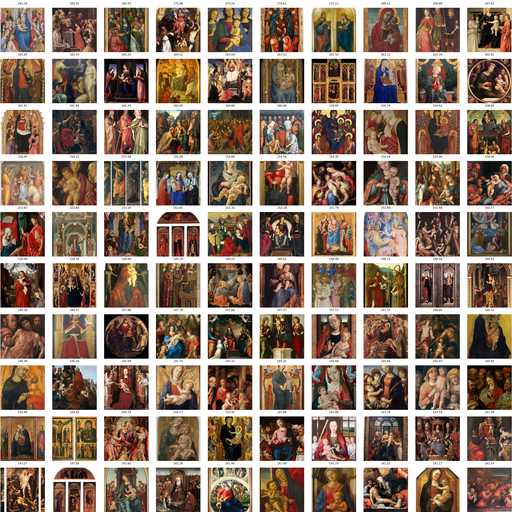}  \caption{} \label{fig:MidLevelFilters_RASTA_4c_78_Top100}  \end{subfigure}
& \begin{subfigure}[c]{0.24\textwidth}       \includegraphics[width=\textwidth]{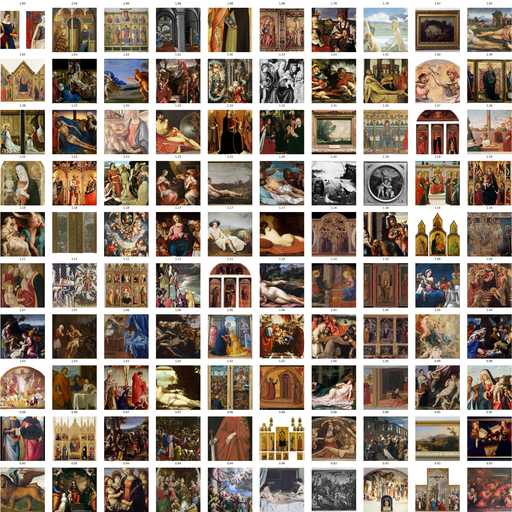}  \caption{} \label{fig:MidLevelFilters_RASTA_4d_49_Top100}  \end{subfigure} \\
\begin{minipage}[c]{0.24\textwidth} \centering  RASTA and IconArt Fine Tuned Maximal Activation Examples   \end{minipage}  &  \begin{subfigure}[c]{0.24\textwidth}     \includegraphics[width=\textwidth]{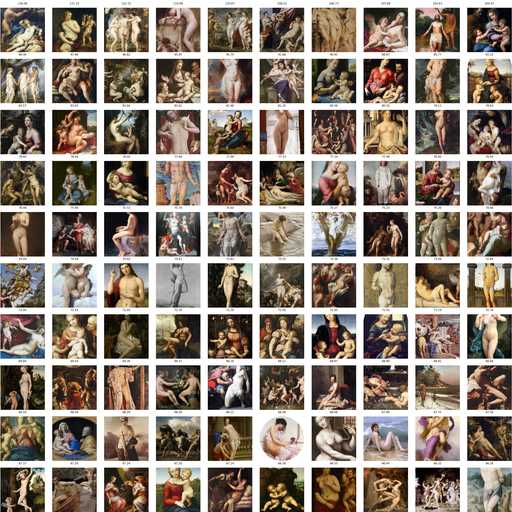} \caption{}  \label{fig:LearnedRASTAIconArt_Top100_nudity} \end{subfigure}
&\begin{subfigure}[c]{0.24\textwidth}      \includegraphics[width=\textwidth]{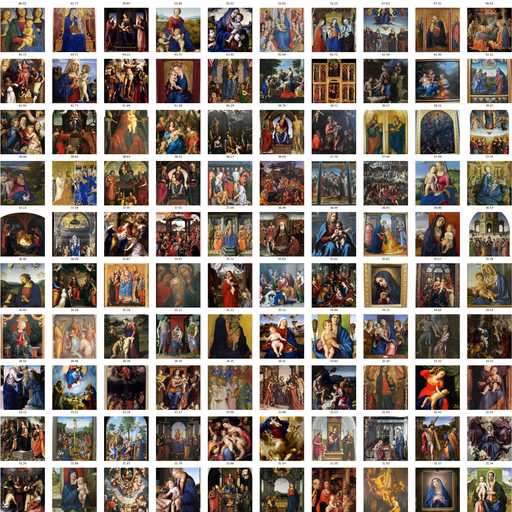} \caption{} \label{fig:LearnedFromScratch_Top100_bluedrapery} \end{subfigure}
& \begin{subfigure}[c]{0.24\textwidth}     \includegraphics[width=\textwidth]{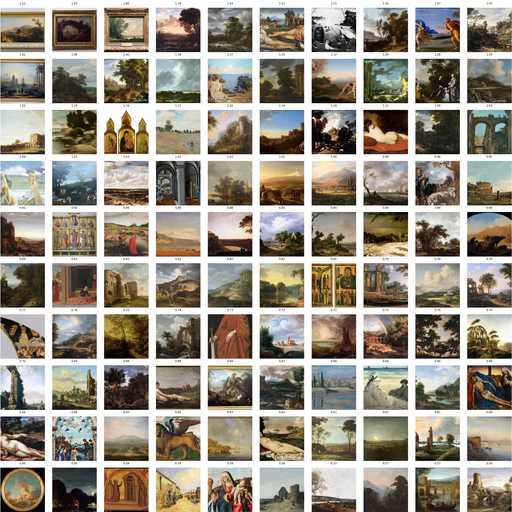}\caption{}  \label{fig:LearnedFromScratch_Top100_treeblue} \end{subfigure} \\ 
    \end{tabular}
     }
     \end{center}
\caption{Maximal Activation Examples (from IconArt train set) the channel corresponding  to \cref{fig:MidLevelFilters_RASTA_IconArt} channels for an InceptionV1 learned on ImagneNet then fine-tuned on RASTA  \protect\cite{lecoutre_recognizing_2017} and finally on IconArt  \protect\cite{gonthier_weakly_2018}.}   
\label{fig:MidLevelFilters_RASTA_IconArt_Top100}
\end{figure} 
\end{subappendices}

\end{document}